\documentclass{article}
\usepackage{arxiv}
\usepackage[utf8]{inputenc}
\usepackage{amsmath,amsfonts,graphicx,color,mathrsfs,bm}

\usepackage{soul}
\usepackage{hyperref}
\usepackage{algorithmic}
\usepackage[ruled]{algorithm2e}
\SetKwInput{KwInput}{Input}   
\SetKwInput{KwOutput}{Output} 
\usepackage{subfigure}

\newcommand{\R}{\mathbb{R}}
\newcommand{\E}{\mathbb{E}}
\newcommand{\A}{\mathscr{A}}
\renewcommand{\d}{\mathrm{d}}
\newcommand{\vect}[1]{\boldsymbol{#1}}
\newcommand{\x}{\vect{x}}
\newcommand{\z}{\vect{z}}

\newcommand{\icol}[1]{
\left(\begin{smallmatrix}#1\end{smallmatrix}\right)%
}

\title{Tipping Points of Evolving Epidemiological Networks:\\ 
Machine Learning-assisted, \\ Data-Driven Effective Modeling}

\author{
  Nikolaos Evangelou\\
  Johns Hopkins University\\
  Baltimore, MD 21218, USA\\
  \And
Tianqi Cui\\
  Johns Hopkins University\\
  Baltimore, MD 21218, USA\\
  \And
  Juan M. Bello-Rivas\\
  Johns Hopkins University\\
  Baltimore, MD 21218, USA\\
  \And
  Alexei Makeev$^{\dag}$ \\
    Moscow State University\\
    Moscow, 119991, Russia \\
  \And
  Ioannis G. Kevrekidis*\\
 Johns Hopkins University\\
  Baltimore, MD 21218, USA\\
  yannisk@jhu.edu
}

\date{May 2022}

\begin{document}

\clearpage

\maketitle

\begin{abstract}
    We study the tipping point collective dynamics of an {\em adaptive} susceptible-infected-susceptible (SIS) epidemiological network in a data-driven, machine learning-assisted manner. 
    We identify a parameter-dependent effective stochastic differential equation (eSDE) in terms of physically meaningful coarse mean-field variables through a deep-learning ResNet architecture inspired by numerical stochastic integrators. 
    We construct an approximate effective bifurcation diagram based on the identified drift term of the eSDE and contrast it with the mean-field SIS model bifurcation diagram. 
   We observe a subcritical Hopf bifurcation in the evolving network's effective SIS dynamics, that causes the tipping point behavior; this
   takes the form of large amplitude collective oscillations that spontaneously -yet rarely- arise from the neighborhood of a (noisy) stationary state. 
   We study the statistics of these rare events both through repeated brute force simulations and by using established mathematical/computational tools exploiting the right-hand-side of the identified SDE. 
   We demonstrate that such a collective SDE can also be identified (and the rare events computations also performed) in terms of {\em data-driven} coarse observables, obtained here via manifold learning techniques, in particular Diffusion Maps. The workflow of our study is straightforwardly applicable to other complex dynamics problems exhibiting tipping point dynamics. 
   
%%%% Here   
%   Furthermore, we perform rare event analysis - mean escape time computation - with our constructed eSDE models and with the full SIS model and discuss the computational benefits of using the eSDE models instead of the full SIS model for these computations. The computation of the escape times was performed: (a) with brute-force stochastic simulations, (b) by solving a stationary boundary value problem (BVP), and (c) by solving a transient initial-boundary value problem (IBVP). %The   transient boundary value problems arising from the Feynman-Kac formula.

%   Our framework can be easily applied to a large number of models but also to experiments to study the tipping point dynamics of those systems in a data-driven manner. 
\end{abstract}

\section{Introduction}

The dependence of dynamical systems on parameters is known to lead to {\em bifurcations}: qualitative changes in the nature of the dynamics (e.g. creation or destruction of fixed points) \cite{strogatz2018nonlinear,guckenheimer2013nonlinear}. 
The critical parameter value at which such a qualitative change occurs is a \textit{bifurcation point}; certain so-called ``hard" bifurcation points (e.g. turning points, subcritical Hopf points) can lead to what has come to be referred to as \textit{tipping point} behavior. Tipping points have been observed across many research areas, from mathematics (e.g. \cite{guckenheimer2013nonlinear}) to engineering (e.g. \cite{uppal1974dynamic}) and from epidemiology (e.g. \cite{gross2006epidemic,gross2008robust}) to weather modeling (e.g. \cite{scheffer2020critical,gnanadesikan2018flux}), economics (e.g. \cite{gualdi2015tipping,liu2015equation,omurtag2006modeling}) and the social sciences (e.g. \cite{centola2018experimental,milkoreit2018defining,otto2020social,gladwell2006tipping}.)

\begin{figure}
    \centering
    \includegraphics[width=7cm]{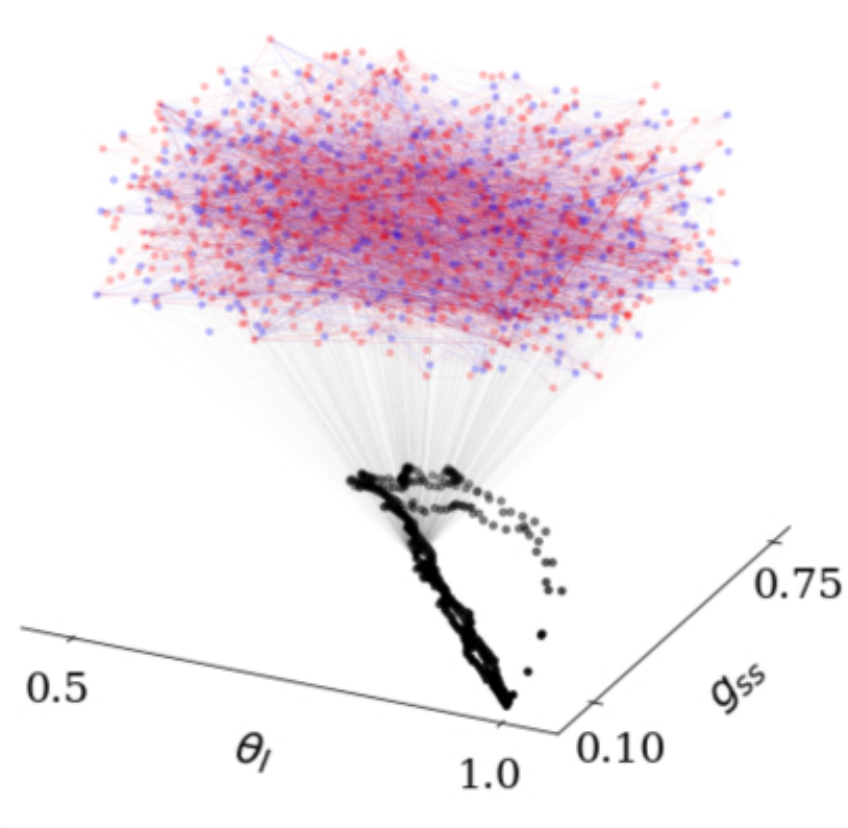}
    \caption{Schematic of the full SIS model. The network's nodes are shown as a colorful cloud of points with connections between them; two different types of nodes, red and blue, denote the infected and susceptible individuals respectively. A projection of the full SIS model's dynamics is shown on the plane of the mean-field coarse variables $\theta_I$ (fraction of infected individuals) and $g_{SS}$ (fraction of edges between susceptible individuals).}
    \label{fig:illustration_SIS}
\end{figure}

In this paper, we focus on the tipping point dynamics observed in an {\em adaptive} epidemiological network model proposed by Gross et al. \cite{gross2006epidemic}. An illustration of the model is shown in Figure \ref{fig:illustration_SIS}. This epidemiological model belongs to the class of Susceptible-Infected-Susceptible (SIS) models but it considers the epidemic dynamics as occurring on an evolving network. Each person in this network appears as a node and can be either Susceptible or Infected. The edges (links) in the network are  seen as social interactions: connections between people. At each (discrete) time step of the full SIS model, each infected individual can recover with probability $r$, and each susceptible individual (through each of its connections to infected individuals) may become infected with probability $p$. The edges (connections) of this network will also change during each time step, because susceptible individuals can exercise the opportunity to avoid the infected individuals they are connected with, by rewiring links to their infected connections with probability $w$, thus making the network \textit{adaptive}. These rules -on a network where both the state of the nodes and the existence of edges- probabilistically evolves  adaptive gives rise to complex dynamics e.g. (noisy) hysteresis and (noisy) large scale oscillatory dynamics \cite{gross2006epidemic,gross2008robust}. 

We begin the study of the dynamics of our networked SIS model by first investigating the dynamics of its simplified mean-field model description, proposed by Gross et al. \cite{gross2006epidemic}, a description that incorporates the rewiring effect (see section \ref{sec:coarse_grained_model}). For this low-dimensional deterministic model, by altering the probability of infection $p$, one identifies a supercritical (soft) Hopf bifurcation for a range of rewiring parameter values. 
To achieve a more detailed understanding of the full stochastic adaptive SIS network dynamics we sample data over a range of different parameter values for the probability of infection - a range where collective oscillations appear.
The observed dynamics -exhibiting sudden jumps to large amplitude collective oscillations- are indicative of an underling {\em hard, subcritical} effective Hopf bifurcation.
We record these tipping point dynamics in terms of coarse mean-field variables computed directly from the network simulation. We identify an {\em effective Stochastic Differential Equation} (eSDE) by using a recently developed deep-learning algorithm, inspired by stochastic numerical integrators, in terms of the mean-field collective variables \cite{dietrich2023learning}. 
We exploit the identified eSDE drift (its deterministic component) to characterize the tipping point dynamics of the model. We construct an approximate effective bifurcation diagram with the AUTO/SCIGMA software \cite{scigma2023}, based on the eSDE drift and pinpoint the critical parameter value at which a {\em subcritical} Hopf Bifurcation occurs. 
We then repeat this workflow, but now using a set of {\em data-driven} observables/variables, obtained by using the manifold learning scheme Diffusion Maps \cite{coifman2006diffusion} (see section \ref{sec:diffusion_maps}). It is important to notice that Diffusion Maps is based on the use of a similarity metric between data points; when the data points are themselves graphs (as is the case in our evolving adaptive network problem) graph metrics must be used to identify good data-driven variables (observables) \cite{aggarwal2010managing,bold2012equation,kattis2016modeling,Karthikeyan2017,wills2020metrics,athreya2022discovering}. We derive such Diffusion Maps data-driven observables of network evolution, identify an eSDE in terms of these observables, and demonstrate how they are related to the (physically interpretable) coarse mean-field variables. 

The two identified eSDE model versions are then used to perform computer-assisted rare-event analysis (escape time computations) for the ``catastrophic jumps" from stationary collective behavio to large amplitude, noisly collective oscillations. We compare the mean escape times of our data-driven reduced dynamical model to the ones from the full epidemics network. The computation of the escape times is performed: (i) through brute force stochastic simulations, (ii) by solving a (steady) boundary value problem \cite{oksendal2013stochastic}
and (iii) by solving an initial-boundary value problem \cite{oksendal2013stochastic}. The differential equations used to formulate the boundary value problems arise from the Feynman-Kac formula formalism \cite{oksendal2013stochastic}. We compare the wall-clock computational time for estimating the escape time distribution for the full epidemiological model with that of the two alternative eSDE surrogate models--  and show that the eSDE models provide significant computational savings.

{\bf Prior Art}  In this subsection we do not focus on the extensive epidemiological literature beyond the references mentioned so far. We rather discuss current techniques for addressing the technical tasks involved in our tipping point computations, and the data science/machine learning techniques that are combined with tradition continuum deterministic modeling software.
A significant number of machine learning techniques for identifying dynamics of ordinary \cite{brunton2016discovering,chen2018neural,NEURIPS2020_4a5876b4,lu2019deeponet}, partial \cite{lu2019deeponet,raissi2019physics,brandstetter2022clifford,kemeth2022learning,lee2022learning}, and stochastic differential equations, from data, has been proposed in the last few years, even though machine learning-assisted identification of dynamical systems started in the 1990s \cite{krischer1993model,rico1993discrete,rico1992discrete,gonzalez1998identification}. Since in this work, we deal with the deep learning identification of stochastic differential equations from data, we provide a more detailed description of recent deep learning approaches in that direction. Li X. et al. \cite{li2020scalable}  proposed an approach for learning an eSDE in latent variables by estimating path-wise gradients of long series.
Generative adversarial networks (GANs) have been also popular tools for learning and/or solving stochastic differential equations. Yang L. et. al \cite{yang2020physics} proposed Physics Informed GANs (PI-GANs) for solving SDEs by encoding governing physical laws in the form of SDEs in the network's architecture. Yang L. et al \cite{yang2022generative} proposed a generative \textit{ensemble-regression} (GER) approach that learns the governing stochastic differential equation from particle densities (ensembles). Kidger P. \cite{kidger2021neural} et al. showed that Wasserstein GANs can be used to fit stochastic differential equations through the generator-discriminator components of the architecture. Zhu Y. et al. \cite{zhu2023learning}  proposed  a framework called the statistics-informed neural network (SINN) based on a long short-term memory (LSTM) architecture that reproduces the statistical characteristics of an SDE given i.i.d. random sequences as input. Fang C. et al. \cite{fang2022end} proposed an approach to identify stochastic differential equations driven by $\alpha$-stable L\'evy noise from pairwise data.
The proposed method by Hasan A. et al \cite{hasan2021identifying} identifies a latent space and the stochastic differential equation in this latent space with a variational auto-encoder. This later method is the most similar to the one we use \cite{dietrich2023learning}. The constructed loss function is based on the numerical scheme Euler-Maruyama (similar to the approach we use) but they consider only isotropic diffusivity.
Our method of choice proposed by Dietrich F. et al. \cite{dietrich2023learning} uses an \textit{effective} ResNet architecture to identify  stochastic differential equations from pairwise data. The constructed loss functions are inspired by numerical stochastic integrators (e.g Euler-Maryuama, Milstein) assuming that the noise is given by a Wiener process. The approximated diffusivity coefficient (in contrast to \cite{hasan2021identifying}) can be non-isotropic and state-dependent. A more detailed description of this approach is given in section \ref{sec:Learning_SDEs}. 

Machine learning and data mining techniques have been proven useful for discovering patterns in high dimensional data and providing reparametrizations of the original data that can lead to dimensionality reduction. Such techniques include Principal Component Analysis \cite{pearson1901liii},  Local Linear Embedding \cite{roweis2000nonlinear}, Laplacian Eigenmaps \cite{belkin2003laplacian}, Isomap \cite{balasubramanian2002isomap}, t-SNE \cite{van2008visualizing}, Umap \cite{mcinnes2018umap}, Autoencoders \cite{kramer1991nonlinear}, and our method of choice, {\em Diffusion Maps} \cite{coifman2006diffusion}. The Diffusion Maps algorithm proposed by Coifman and Lafon \cite{coifman2006diffusion} is a non-linear dimensionality reduction technique that has been used in a broad range of applications to discover intrisic latent observables (variables or even parameters) \cite{evangelou2022double,evangelou2021parameter,koronaki2023partial,talmon2013diffusion,nadler2006diffusion,holiday2019manifold,sroczynski2022questionnaires,yair2017reconstruction,chiavazzo2017intrinsic,ferguson2010systematic,evangelou2022learning,kemeth2021initializing}. In our work, Diffusion Maps was applied to data sampled from the full SIS adaptive evolving network to parametrize the \textit{intrinsic} geometry of the data and discover a set of latent coarse (collective, effective) data-driven variables. A more detailed description of the Diffusion Maps algorithm is provided in section \ref{sec:diffusion_maps}. The adaptation of Diffusion Maps to the case where its data point studied is itself a graph has been studied in e.g. \cite{bold2012equation,kattis2016modeling,Karthikeyan2017}.

To study, predict, and anticipate tipping point dynamics, researchers have developed data-assisted (e.g. the \textit{equation-free} approach \cite{kevrekidis2003equation} ) and machine learning techniques over the last 30 years or so. In the early 90s Rico-Mart\'inez R. et. al \cite{rico1993discrete} used neural network architectures inspired by explicit and implicit numerical integrators for ordinary differential equations to construct bifurcation diagrams based on experimental data. The equation-free approach developed in the early 2000s by our group has been successfully applied to the construction of coarse-grained bifurcation diagrams. 
This approach performs selected short bursts simulations of fine-scale models, constructs {\em local} surrogate models in terms of known coarse variables, and uses those in a data-assisted fashion to perform the continuation in the construction of the bifurcation diagrams. 
More recently, Bury M. T. et al. \cite{bury2021deep} used deep learning algorithms (convolutional neural networks and LSTMs) to build a classifier that detects early warning signals of tipping points for fold, Hopf, and transcritical bifurcations. In the last few years, reservoir computing has been used for the prediction-anticipation of critical transitions for stationary ODEs, non-stationary ODEs, stochastic differential equations, and partial differential equations \cite{patel2023using,lim2020predicting,kong2021machine,xiao2021predicting}. In those works, the authors illustrated that reservoir computing is capable of predicting the occurrence of a tipping point and in some cases also predicting the post-tipping point dynamics behavior. In contrast to the previous approaches, that use reservoir computing to illustrate that machine learning algorithms can be used to anticipate tipping point dynamics, Galaris et. al \cite{galaris2022numerical} constructed the bifurcation diagram of a ``black-box'' coarse-grained PDE given sampled data over the parametric space of a finer scale Lattice Boltzmann model. In addition, Fabiani et al. \cite{fabiani2023tasks} proposed a  machine learning (ML)-assisted framework for constructing the bifurcation diagram for stochastic agent-based models either with identified mesoscopic  Integro-Partial Differential Equations (IPDEs) and mean-field-type effective SDEs constructed. 

In this work we depart from the equation-free approach, since our constructed surrogates are not \textit{local}. We identify eSDEs in terms of the coarse (mean-field and data-driven) variables, and we use those surrogates for rare event analysis and construction of bifurcation diagrams. In contrast to previous work in the literature \cite{patel2023using,lim2020predicting,kong2021machine,xiao2021predicting}, our eSDE is not used for early warning signals of tipping points but rather to disentangle the contribution of the deterministic dynamics (drift term) from the stochastic dynamics (diffusivity term). This allows us to (i) utilize the identified \textit{effective-ODE} dynamics from the eSDE so to construct the bifurcation diagram, and to (ii) study a meaningful ``skeleton" of the dynamic behavior in terms of the coarse variables discarding the noise. %Furthermore, the constructed data-driven eSDE models are used for escape time computations and the obtained results are contrasted with the full SIS model. We show a reduction in computational time compared to the full SIS model. 
 
The remainder of this paper is organized as follows: In Section \ref{sec:SIS_Model}, we provide a detailed description of the \textit{rules} that govern the full SIS network  model. In section \ref{sec:coarse_grained_model}, we study the low-dimensional mean-field model of Gross et al. \cite{gross2006epidemic} and discuss some of its differences from the full SIS network model. 
In section \ref{sec:diffusion_maps}, we provide details for the Diffusion Maps algorithm and mention how it can be applied when each data point is a snapshot of an evolving graph. In section \ref{sec:Learning_SDEs}, we explain the deep-learning algorithm used for the identification of stochastic differential equations. In section \ref{sec:escape_times_methodology} we describe the different approaches we implemented for the computation of the escape times. In section \ref{sec:physical_coarse_variables}, we illustrate sampled trajectories across two different parameter values in terms of physically meaningful coarse variables, and provide a qualitative comparison of the resulting types of behavior. In sections \ref{sec:learned_SDEs}
and \ref{sec:bifurcation_diagrams_comparison} we illustrate the identified attractors based on the drift for each of the two parameter values, and provide a comparison of the constructed bifurcation diagrams between the mean-field model and the one identified by the eSDE in terms of the mean-field variables. In section \ref{sec:diffusion_maps_results} we show the data-driven variables obtained with Diffusion Maps, discuss their relation to physical mean-field variables, and illustrate that the identified eSDE on those gives the same (drift-deterministic) attractor as the one estimated using the physical mean-field variables. In section, \ref{sec:Mean_Escape_Time_results} we provide a comparison between the obtained statistics for the mean escape time values across the three different models (two eSDE models and the full SIS model) and the different approaches we used. In section \ref{sec:computational_savings} we discuss the computational benefits of using the eSDE models --instead of the full SIS model-- for the computation of escape times. Finally, in section \ref{sec:discussion_conclusion}, we summarize our approach and offer our views of its potential, shortcomings, current and future research directions.

\section{Methodology}

\subsection{The full SIS adaptive network epidemiological model}
\label{sec:SIS_Model}
The adaptive susceptible-infected-susceptible (SIS) model could be considered as a labeled undirected evolving graph, $G$. This labeled graph has fixed number of nodes $N$  and fixed number of links $L$ (edges). In the simulations reported below we chose $N= 10,000$ and $L= 10N$. Each node representing an individual can be in two states, either infected (I) or susceptible (S). The edges between nodes (social connections) are defined based on the state of the nodes they connect as susceptible-susceptible (SS-links), infected-infected (II-links), and susceptible-infected (SI-links). In every (discrete) time step of the model 3 substeps are taking place:
\begin{enumerate}
    \item The infected individuals with a probability $r$ recover and they become susceptible.
    \item Each susceptible connected with an infected individual (SI-link) becomes infected with probability $p$.
    \item The connection between each susceptible connected to an infected individual (SI-link) is removed with probability $w= w_0\theta_I = w_0\frac{i}{N}$ and a \textit{new link} from this susceptible node\textit{ to another randomly selected susceptible node} is created. The parameter $w_0$ is constant but  $\theta_I=\frac{i}{N}$, the fraction of infected individuals is not, since it depends on the number of infected individuals $i$ in the network at a given time.
\end{enumerate}

The adaptive rewiring parameter $w$ allows the susceptible individuals to avoid contact with infected individuals. The probability of avoiding infected individuals increases when $\theta_{I}$ increases. This assumes that the information of the disease (large number of $\theta_I$) is instantaneously transmitted across the network and the susceptible individuals become more \textit{cautious} and avoid infected individuals.

\begin{figure}[ht]
    \centering
    \includegraphics[width=16cm]{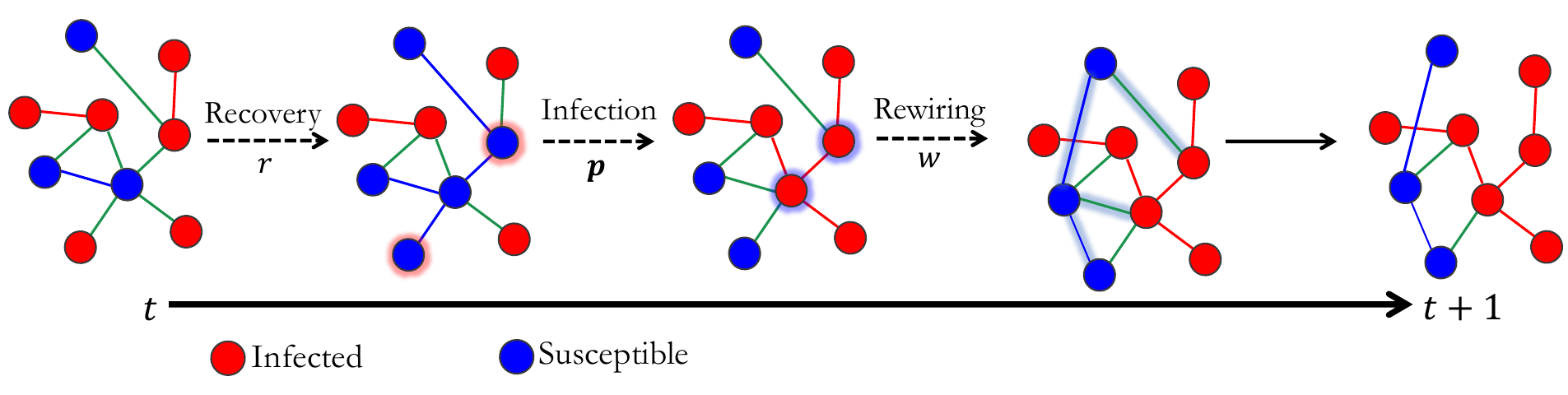}
    \caption{A schematic for one time step of the SIS model. The nodes representing infected individuals (I) are shown with red color. The nodes representing susceptible individuals (S) are illustrated with blue color. The II-links are shown as red edges, the SS-links are shown as blue edges and the SI-links as green edges. Given an initial graph, the first substep involves the recovery of the infected individuals with probability $r$, making them susceptible. During the second substep, the susceptible individuals connected to infected individuals (SI-links) are being infected with probability $p$. The third substep involves the \textit{rewiring of the edges} between the susceptible individuals with infected ones (SI-links) with probability $w$.}
    \label{fig:sis_model_evolution}
\end{figure}

One step of the SIS model's evolution is shown in figure \ref{fig:sis_model_evolution}.

\subsection{The mean-field SIS model}
\label{sec:coarse_grained_model}
To investigate how the rewiring \textit{w} affects the dynamics of the adaptive SIS epidemiological model Gross et al. \cite{gross2006epidemic} considered the following analytical model in terms of the mean-field quantities $i, l_{SS}$ and $l_{II}$

\begin{equation}
    \begin{aligned}
        & \frac{d}{dt} i = pl_{SI} - ri \\
        & \frac{d}{dt} l_{II} = pl_{SI}\bigg(\frac{l_{SI}}{s} + 1\bigg) -2rl_{SS} \\
        &  \frac{d}{dt}l_{SS} = (r+w')l_{SI} - \frac{2pl_{SI}l_{SS}}{s}.
    \end{aligned}
    \label{eq:equations_Gross}
\end{equation}
In Equation \eqref{eq:equations_Gross}, $i$ is the number of infected individuals; $l_{SI}$  represents the number of edges (links) between infected and susceptible individuals; $l_{II}$ represents the number of edges between infected individuals; $l_{SS}$ represents the number of edges between susceptible individuals, and $s$ is the number of susceptible individuals. The number of edges between susceptibles  and infected ($l_{SI}$) and the number of susceptibles $s$ are computed from the conservation laws
\hbox{$L = l_{II} + l_{SI} + l_{SS}$} and $N  = s + i$. For our computations the parameters above were selected as $N=10^5, L = 10^6, r = 0.002 $
following \cite{gross2006epidemic}.

The parameter that governs the probability of recovering $r$ and the probability of infection $p$ in this simple system of equations are analogous to those for the full network; but the probability of rewiring $w'$ constant for this simple model, does not depend on the number of infected individuals as in the full SIS network model. 
As rewiring occurs ($w'>0$), discontinuous transitions, bistability and hysteresis loops arise in the dynamics. In section \ref{sec:bifurcation_diagrams_comparison} we present the bifurcation diagram w.r.t. to $p$ obtained with AUTO for the system of equations \eqref{eq:equations_Gross} at $w' = 0.6$. We also report in section \ref{sec:Bifurcation_Coarse_ODEs} of the SI the bifurcation diagrams obtained with AUTO for this system of equations \eqref{eq:equations_Gross} for $w' = \{0.2,0.4,0.6\}$ \cite{gross2006epidemic,gross2008robust}.

\subsection{Diffusion Maps for data in the form of networks}
\label{sec:diffusion_maps}

Diffusion Maps, introduced by Coifman and Lafon \cite{coifman2006diffusion,coifman2005geometric}, can reveal the \textit{intrinsic geometry} of a data set \hbox{$\mathbf{X} = \{\vect{x}\}_{i=1}^m$}, where the data points $\vect{x}_i \in \mathbb{R}^d$ are sampled from a manifold $\mathcal{M}$. In this section, we describe how the Diffusion Maps algorithm applied to vector data can achieve dimensionality reduction; later we discuss how Diffusion Maps can be extended for data sets where each data point is a graph. The Diffusion Maps algorithm constructs a random walk on $\mathbf{X}$ by computing an affinity matrix $\mathbf{A} \in \mathbb{R}^{m \times m}$ that measures the local similarity between the data points $\vect{x}_i$. The entries of $\mathbf{A}$ are computed with a kernel function. In our case the Gaussian kernel,

\begin{equation}
    A(\vect{x}_i,\vect{x}_j) = \exp\bigg( \frac{-||\vect{x}_i - \vect{x}_j ||^2}{2\varepsilon} \bigg),
\end{equation}
 was used where $\varepsilon$ is a positive scale hyperparameter that regulates the rate with which the kernel decays and $||\cdot||$ denotes the norm of choice, typically the $\ell^2$ norm. A normalization applied to $\mathbf{A}$ provides a parametrization of the data regardless of the sampling density. This is achieved by first constructing the diagonal matrix
\begin{equation}
    P_{ii}  = \sum_{j=1}^mA_{jj}
\end{equation}
and then computing $ \widetilde{\mathbf{A}}$ as
\begin{equation}
    \widetilde{\mathbf{A}} = \mathbf{P}^{-1}\mathbf{A}\mathbf{P}^{-1}.
\end{equation}
The normalization

\begin{equation}
    W(\vect{x}_i,\vect{x}_j) = \frac{\widetilde{A}(\vect{x}_i,\vect{x}_j)}{\sum_{j=1}^m \widetilde{A}(\vect{x}_i,\vect{x}_j)},
\end{equation}

is then applied on $\widetilde{\mathbf{A}}$ yielding the Markov Matrix $\mathbf{W}$. The eigendecomposition of $\mathbf{W}$,
\begin{equation}
    \mathbf{W}\vect{\phi}_i = \lambda_i\vect{\phi}_i
\end{equation}
provides a set of eigenvectors $\vect{\phi}_i$ and eigenvalues $\lambda_i$. Selecting the eigenvectors $\vect{\phi}_i$ that span unique directions, termed \textit{non-harmonic eigenvectors}, in the data set, provides a reparametrization of $\mathbf{X}$ in terms of the \textit{Diffusion Maps coordinates}. The selection of the non-harmonic eigenvectors can be performed by visual inspection if the dimensionality is smaller or equal to three, or by the local linear regression algorithms that systematically select those eigenvectors \cite{evangelou2022double,dsilva2018parsimonious}. If the number of non-harmonic eigenvectors is smaller than $d$ then dimensionality reduction has been achieved.

\begin{figure}[ht]
    \centering
    \includegraphics[width=8cm]{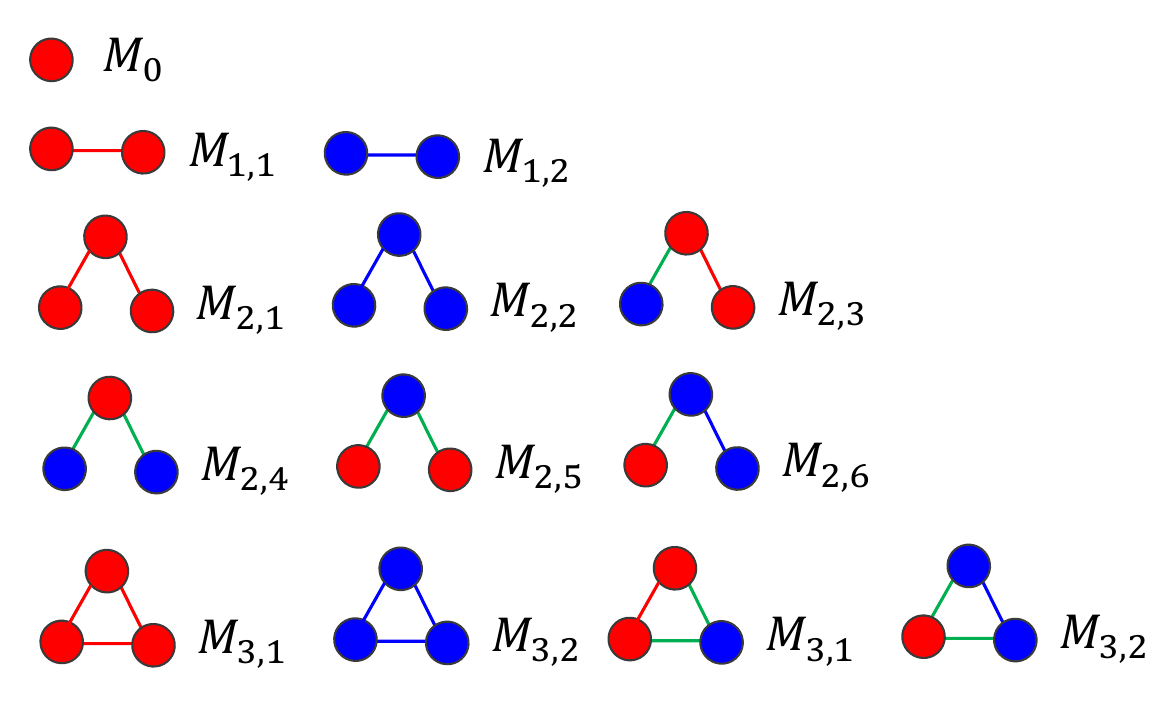}
    \caption{The 13 motifs/subgraphs enumerated from each graph towards the Diffusion Maps computation.}
    \label{fig:motifs_}
\end{figure}

In our case, where each data point is not a vector ($\vect{x}_i$), but a graph $G_i$, a metric/similarity measure needs to be defined across graphs. Based on our previous work, \cite{kattis2016modeling,Karthikeyan2017} we considered a \textit{motif} distance for our labeled graphs. We chose 13 subgraphs/motifs, shown in figure \ref{fig:motifs_}, and for each graph $G_i$ sampled in our data set, we enumerated the number of those 13 motifs; we also considered normalized motif fractions. From this computation, each graph maps to a 13-dimensional vector of motifs $\vect{s}_i \in \mathbb{R}^{13}$. The Diffusion Maps algorithm, presented above in the more general case, is then applied in the data set of the collected enumerated motifs $\mathbf{S} = \{\vect{s}_i\}_{i=1}^m$. These 13 motifs, shown in figure \ref{fig:motifs_} provide a \textit{summary} of the intrinsic information for each graph
including the two different states (susceptible and infected). Phase portraits of the network SIS evolution in such graph Diffusion Maps variables will be presented below.

It is important to argue why a lower dimensional behavior/structure may arise for those large adaptive networks. In principle, large adaptive networks can have millions of degrees of freedom. However, most possible network configurations will probably not appear in the \textit{long-term dynamics}. This suggests that a time-scale separation may exist between a few, slow, collective variables and the remaining fast, fine-scale variables; and that the long-term dynamics \textit{live} in an effectively low-dimensional manifold parametrized by the few slow variables. This lower-dimensional manifold in our case is discovered by applying the Diffusion Maps algorithm discussed above \cite{gross2008robust,kattis2016modeling}

\subsection{Learning stochastic differential equations}
\label{sec:Learning_SDEs}
\begin{figure}[ht]
    \centering
    \includegraphics[width=14cm]{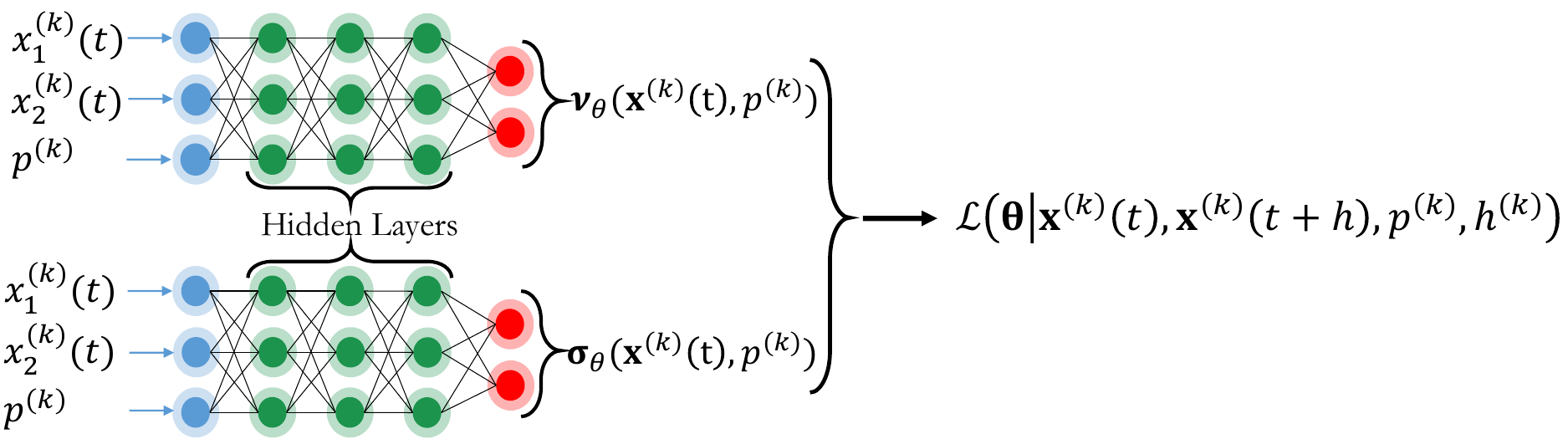}
    \caption{A schematic of the deep learning framework we used to estimate the eSDE through the two neural networks $\vect{\nu}_{\theta}$ and  $\vect{\sigma}_{\theta}$ for the two-dimensional case $\vect{x}^{(k)}(t) = \icol{x_1^{(k)}(t)\\x_2^{(k)}(t)}$. The two neural networks are being trained simultaneously and are connected in terms of the loss function $\mathcal{L}$. }
    \label{fig:SDE_NN}
\end{figure}

The Stochastic Differential Equations we consider in this work is of the Langevin form
\begin{equation}
\label{eq:sde_langevin}
    d\vect{x}(t) = \vect{\nu}(\vect{x}(t);p)dt + \bm{\sigma}(\vect{x}(t);p)d\vect{B}_t 
\end{equation}
where $\vect{x}(t) \in \mathbb{R}^{n}$ is a vector of stochastic variables, $\vect{\nu}: \mathbb{R}^{n} \mapsto \mathbb{R}^{n}$ is the drift, $\bm{\sigma}: \mathbb{R}^n \mapsto \mathbb{R}^{n \times n}$ is the diffusivity matrix, $p$ is the parameter that affects the dynamics, in our case, $p$ is the probability of infection discussed in section \ref{sec:SIS_Model}, and $\vect{B}$ is a collection of $n$ one-dimensional Wiener processes. The dynamics of an SDE of the above form (Equation \eqref{eq:sde_langevin}) can be approximated by estimation of the drift ($\vect{\nu}$) and diffusivity ($\bm{\sigma}$) coefficients from data. The approximation of the two coefficients (drift and diffusivity) in our work was made based on the methodology proposed by Dietrich et al \cite{dietrich2023learning} (and the modification introduced by Evangelou et al \cite{evangelou2022learning}) that allows identifying a parameter-dependent eSDE. The proposed methodology estimates the drift and diffusivity through two neural networks $\vect{\nu}_{\theta}$ and $\sigma_{\theta}$ where $\theta$ are the weights of the networks. The two networks are connected through the loss function, as shown in figure \ref{fig:SDE_NN} for a two-dimensional eSDE.

To train this network we assume that we have access to $m$ snapshots of the form \hbox{$D = \{\vect{x}^{(k)}(t+h),\vect{x}^{(k)}(t),h^{(k)},p^{(k)} \}_{i=1}^m$ }, where $\vect{x}^{(k)}(t)$ is a data point sampled at time $t$, $\vect{x}^{(k)}(t+h)$ is the evolution of $\vect{x}^{(k)}(t)$ after a short time $h^{(k)}>0$, $p^{(k)}$ is the parameter that affects the dynamics of Equation \eqref{eq:sde_langevin}. Those snapshots in our case were extracted from long simulated trajectories of the model described in section \ref{sec:physical_coarse_variables}. 

 The loss functions proposed by Dietrich et al. \cite{dietrich2023learning} are derived based on numerical stochastic differential equations (e.g. Euler-Maruyama, Milstein). In this work, the Euler-Maruyama loss function template was used and is described below for the 2D case, $\vect{x}^{(k)}(t) = \icol{x_1^{(k)}(t)\\x_2^{(k)}(t)}$.

 The Euler-Maruyama numerical integration scheme of Equation \eqref{eq:sde_langevin} for a time step $h^{(k)}$ gives

 \begin{multline}
 \label{eq:Euler_Maruyamar}
\begin{pmatrix}
 x_1^{(k)}\left(t+h^{(k)}\right) \\
  x_2^{(k)}\left(t+h^{(k)}\right)
\end{pmatrix}
=
\begin{pmatrix}
 x_1^{(k)}(t) \\
  x_2^{(k)}(t)
\end{pmatrix}
+
h^{(k)}
\begin{pmatrix}
 \nu_{\theta ,1}(\vect{x}^{(k)}(t);p^{(k)}) \\
   \nu_{\theta ,2}(\vect{x}^{(k)}(t);p^{(k)})
 \end{pmatrix}
 + 
\begin{pmatrix}
  \sigma_{\theta,1}(\vect{x}^{(k)}(t);p^{(k)}) & 0 \\
   0 &
   \sigma_{\theta,2}(\vect{x}^{(k)}(t);p^{(k)})
 \end{pmatrix}
 \begin{pmatrix}
  \delta B_{t_1} \\
   \delta B_{t_2}
 \end{pmatrix}
\end{multline}

where $\nu_{\theta ,1}(\vect{x}^{(k)}(t);p^{(k)})$ is the first component of the drift and $\sigma_{\theta,1}(\vect{x}^{(k)}(t);p^{(k)})$ is the first diagonal component for the diffusivity, $\delta B_{t_1}$ and $\delta B_{t_2}$ are normally distributed with mean zero and variance $h^{(k)}$. The deep-learning scheme proposed by Dietrich et al. \cite{dietrich2023learning} is capable of identifying also off-diagonal terms of the diffusivity matrix $\bm{\sigma}$ but in this work our computations were peformed assuming that the off-diagonal terms were zero.

Equation \eqref{eq:Euler_Maruyamar} implies that we can draw each $\vect{x}^{(k)}(t+h)$ from a normal distribution of the form

\begin{equation}
\label{eq:normal_distribution}
    \vect{x}^{(k)}(t+h) \sim \mathcal{N}\big(\vect{x}^{(k)}(t) + h^{(k)}\nu_{\theta}(\vect{x}^{(k)}(t);p^{(k)}), h^{(k)}\bm{\sigma}(\vect{x}^{(k)}(t);p^{(k)})\big)
\end{equation}

 where the mean is $\vect{\mu}_{\theta}$ of the distribution $\vect{\mu}_{\theta}^{(k)} = \vect{x}^{(k)}(t) + h^{(k)}\vect{\nu}_{\theta}(\vect{x}^{(k)}(t);p^{(k)}) $ and the covariance is \hbox{$\bm{\Sigma}_{\theta}^{(k)}  =   h^{(k)}\bm{\sigma}(\vect{x}^{(k)}(t);p^{(k)})$}. Under the assumption that the next time step data conform with a normal distribution of form shown in Equation\eqref{eq:normal_distribution}, we can maximize the log-likelihood estimation given the data, and thus approximate the drift $\vect{\nu}_{\theta}$ and diffusivity  $\vect{\sigma}_{\theta}$ from the two neural networks. The logarithm of the normal distribution (Equation \eqref{eq:normal_distribution}) and the loss we optimize during training
 is given by the relation

 \begin{multline}
    \mathcal{L}(\theta|\vect{x}^{(k)}(t+h),\vect{x}^{(k)}(t),h^{(k)},p^{(k)}) := 
\log\big|\text{det}(\vect{\Sigma}_{\theta}^{(k)})\big| + \frac{1}{2}(\vect{x}^{(k)}(t+h) - \vect{\mu}_{\theta}^{(k)})^\text{T}\vect{(\Sigma}_{\theta}^{(k)})^{-1}(\vect{x}^{(k)}(t+h) - \vect{\mu}_{\theta}^{(k)}).
\end{multline}
The constant term $\log(2\pi)$ is omitted since it does not affect the optimization. The ability to use two networks to identify the drift and diffusivity is important in our case, since it allows us to \textit{disentangle} the deterministic component (drift) from the stochastic component (diffusivity). By disentangling the identified deterministic dynamics from the stochastic dynamics, we can observe the \textit{on-average} behavior of the system (e.g. the expected value of the upcoming state) and perform further scientific computations (e.g. construct an approximate bifurcation diagram) based only on this.

\subsection{Escape time computation}
\label{sec:escape_times_methodology}

\begin{figure}[ht]
    \centering
\includegraphics[trim={0cm 2.25cm 0cm 3cm},clip,width=0.4 \textwidth]{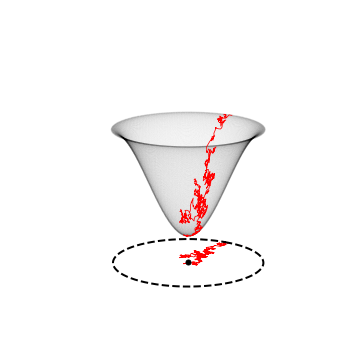}
    \caption{An illustrative example for a stochastic trajectory starting from the initial condition shown with a black solid circle escaping from the unstable limit cycle (dashed lines). The effective energy landscape is shown as a surface above the unstable limit cycle.}
    \label{fig:caricature_escape_time}
\end{figure}

The computation of escape times from a metastable stationary state can be performed with brute force stochastic simulations, as well as by solving boundary value problems (BVPs) as discussed here. For the computation of escape times, we consider the case of a stochastic differential equation (SDE) in the form of Equation \eqref{eq:sde_langevin}. The deterministic attractor, based on the drift component, possesses a stable steady state surrounded by an unstable limit cycle. We estimate the time that it takes for stochastic trajectories starting from the stationary state $\vect{x}_{t_0} = \vect{x}(t=0)$ to escape from the unstable limit cycle for the first time. To help the reader understand the main idea here, in figure \ref{fig:caricature_escape_time}, an illustration is shown where a stochastic trajectory initiated in the stationary state (indicated with a black solid circle) evolves in time until the first time it \textit{escapes} from the unstable limit cycle. The effective energy landscape is shown as a surface above the unstable limit cycle to convey more clearly the concept of \textit{escaping}. Because the nature of the dynamics we are examining is stochastic (Equation \eqref{eq:sde_langevin}) it is important to simulate multiple trajectories to obtain a good averaged estimation of the escape time. In Algorithm \ref{alg:kinetic_Monte_Carlo} the steps for the escape time computations with brute force stochastic simulations are described. The simulation of those trajectories can be performed with any numerical SDE integration scheme of choice. The Euler–Maruyama integration scheme for SDEs was used for our computations (as shown also in Algorithm \ref{alg:kinetic_Monte_Carlo}). To check when $\vect{x}(t_{j+1}) \notin \Omega$ a parameterization of the boundary $\partial\Omega$ is necessary. In our case, the parametrization of the boundary - unstable limit cycle - was performed with the \textit{shapely} Python library \cite{gillies2013shapely}.

\begin{algorithm}[ht]
\caption{Mean escape time computation with brute force stochastic simulations}
  \KwInput{number of simulations $N$, boundary set $\partial \Omega$, initial condition $\vect{x}_{t_0}= \vect{x}(t=0)$, time step $h$}
    \KwResult{Mean escape time $\overline{\tau}$}
\begin{algorithmic}
\FOR{$i = 1$ to $N$}
\STATE Start a simulation at $\vect{x}_{t_0}$
    \STATE $j \gets 0$
    \WHILE{$\vect{x}(t_{j+1}) \in \Omega$}
        \STATE $\tau^i \gets t_j + h$
        \STATE $\vect{x}(t_{j+1}) \gets \vect{x}(t_j) + h\vect{\nu}(\vect{x}(t_j)) + \vect{\sigma}(\vect{x}(t_j)) \delta \vect{B}_t$
\ENDWHILE
\STATE $\mathcal{S}^{\tau} \cup \{\tau^i\}$
\ENDFOR
\STATE $\overline{\tau} = \frac{1}{N}\sum_{i =1}^N \tau^{i}$
\end{algorithmic}
\label{alg:kinetic_Monte_Carlo}
\end{algorithm}

We also pursued two alternative approaches to estimating escape times;
namely, the solution of a boundary value problem (involving an
elliptic PDE) that we refer to as \textit{stationary BVP} and an
initial-boundary value problem (involving a parabolic PDE) that we refer to as \textit{transient IBVP}.
A brief description of these approaches of the BVP and IBVP approaches can be found in section \ref{sec:derivation_BVP_IBVP}.
In our work, the above computations were performed using the finite element method, as implemented by the FEniCSx library \cite{logg2010dolfin,logg2012automated}.

% In section \ref{sec:Circle_} of the SI we present the escape time computation with each of the aforementioned approaches for a \textit{toy} example. For that example, the escape time can be computed analytically and thus it it validates our numerical approaches.

\section{Results}
\subsection{Sampled trajectories of the full SIS network model}
\label{sec:physical_coarse_variables}

We start with simulations of the full SIS adaptive network model over a range of values of the infectiousness parameter $p$. In this section, we illustrate trajectories for $p=\{7.3 \times 10^{-4}, 7.5 \times 10^{-4}\}$ (probability of infection) for random initial conditions. Trajectories for a larger range of parameters are reported in section \ref{sec:Sampling} of the Appendix. Figures  \ref{fig:Sampled_Trajectories_a} and \ref{fig:Sampled_Trajectories_b} show the phase portrait and representative trajectories in terms of the fraction of the infected individuals $\theta_I$ and the fraction of the edges (social links) between individuals $g_{SS}$ for $p=7.3 \times 10^{-4}$ and  $p= 7.5 \times 10^{-4}$ respectively. 
%
%Sampled trajectories from the full epidemiological networks are shown in terms of the two physical coarse variables fraction of the infected individuals $\theta_I$ and fraction of the edges between susceptible individuals $g_{SS}$. 
%
Despite their stochasticity, it becomes visually evident that for both values of the parameter the trajectories show an oscillatory behavior. However, the trajectory for $p = 7.5 \times 10^{-4}$ has both small and larger oscillations, in contrast to the one at $p=7.3 \times 10^{-4}$ for which only large oscillations prevail. This qualitative difference in the behavior of the sampled trajectories might serve as a first indication that a change in the models' behavior occurred because some underlying bifurcation happened.

\begin{figure}[ht]
    \centering
    \subfigure[]{\includegraphics[trim={0.0cm 0.0cm 0.0cm 1.5cm},clip,width=8cm]{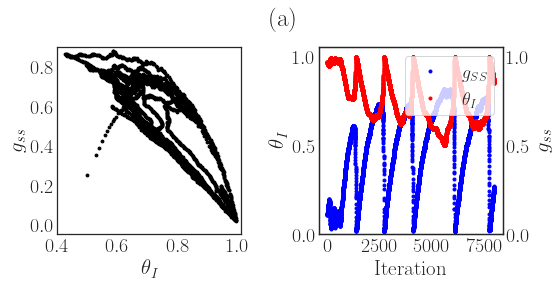}
    \label{fig:Sampled_Trajectories_a}}
    \subfigure[]{\includegraphics[trim={0.0cm 0.0cm 0.0cm 1.5cm},clip,width=8cm]{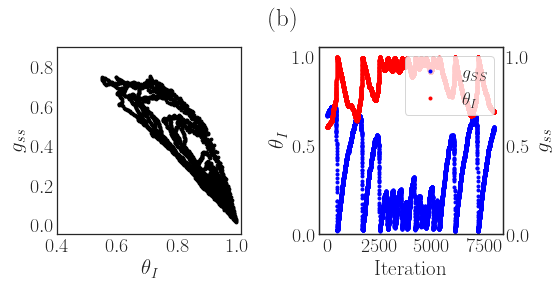}
    \label{fig:Sampled_Trajectories_b}}
\caption{Sampled trajectories from the full epidemiological network projected in the coarse physical variables $\theta_I$ and $g_{SS}$ for (a) $p = 7.3 \times 10^{-4}$ (b) for $p=7.5 \times 10^{-4}$.}
%\label{fig:Sampled_Trajectories}
\end{figure}

\subsection{Learning an eSDE on physical coarse variables}

\label{sec:learned_SDEs}
\begin{figure}[ht]
     \centering
         \includegraphics[width=0.65\textwidth]{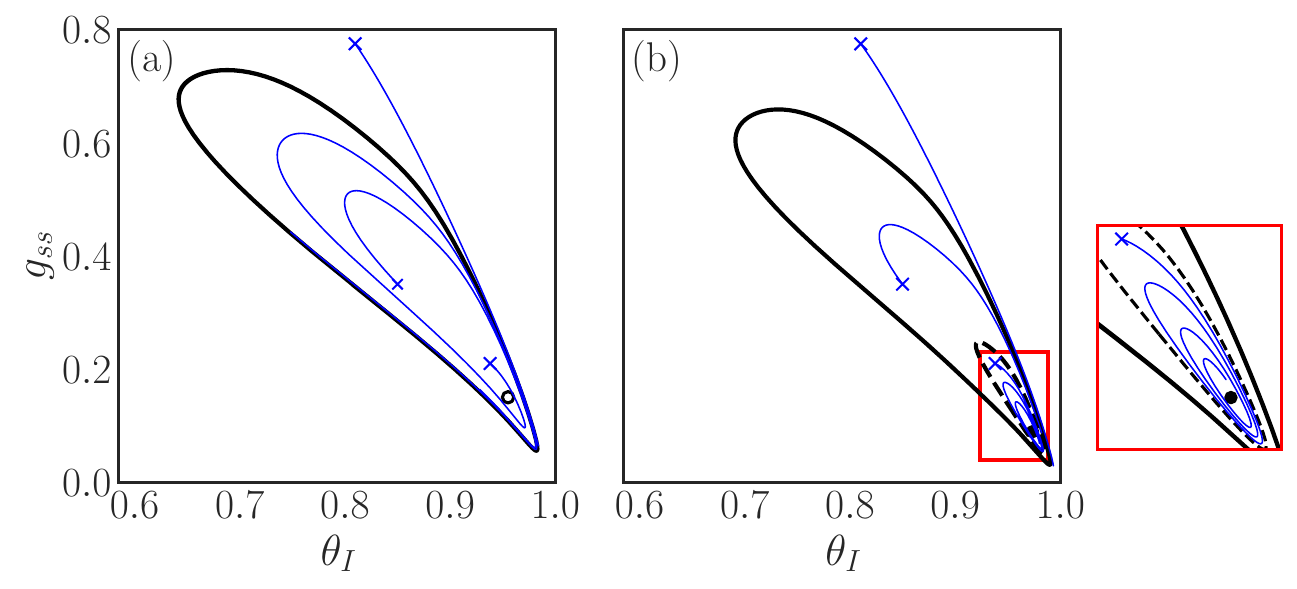}
\label{fig:High Dimensonal Trajectory Error}    
\caption{(a) For $p=7.3 \times 10^{-4}$ the deterministic component of the eSDE dynamics predicts a stable limit cycle (solid black line) with an unstable steady state (white circle) inside it. (b) For $p=7.5 \times 10^{-4}$ the same eSDE drift dynamics predict both a stable limit cycle (solid black line) \textit{and} a stable fixed point (solid black circle). Between them,\textit{ an unstable limit cycle (dashed black line) is predicted}. For both values of the parameter $p$ in (a) and (b), the blue solid lines illustrate trajectories with initial conditions shown with blue crosses.}
\label{fig:deterministic_dynamics}    
\end{figure}

To further investigate the change in the behavior of the dynamics, shown in figure \ref{fig:Sampled_Trajectories_a} and \ref{fig:Sampled_Trajectories_b} we trained a neural network based on the framework described in section \ref{sec:Learning_SDEs} that identifies a parameter-dependent, state-dependent eSDE. For the neural network’s training, we used successive snapshot pairs of $\theta_I, g_{SS}$ sampled from the full SIS model. Sub-sampling of the original sampled data was used by selecting snapshots every five iterations of the full model. To train the eSDE model we scaled iteration steps to a step size $h=0.01$ between the snapshots. The selection of step-size $h$ was considered as a hyperparameter during the neural network's training.  More details of the model's training and its architecture can be found in section \ref{sec:neural_network_details} of the Appendix. The identified eSDE allow us to \textit{disentangle} (separate) the deterministic component (drift) from the stochastic component (diffusivity) and study the system's expected behavior. For the two sampled parameter values the identified dynamics \textit{based on the drift of the eSDE only} are shown in figure \ref{fig:deterministic_dynamics}. The identified deterministic dynamics from the eSDE for $p=7.3 \times 10^{-4}$ illustrate the existence of an unstable steady state surrounded by a stable limit cycle. The identified determinstic dynamics for $p=7.5 \times 10^{-4}$ illustrate the existence of a stable steady state surrounded by an unstable limit cycle. The unstable limit cycle is itself surrounded by a larger, stable limit cycle. The change in stability of the steady state and the creation of an unstable limit cycle at $p = 7.5 \times 10^{-4}$ indicates that a subcritical Hopf bifurcation occurs for a parameter value between our two samples. 

\subsection{Comparison of predicted bifurcation diagrams}
\label{sec:bifurcation_diagrams_comparison}
\begin{figure}[ht]
    \centering
\subfigure[]{   \includegraphics[width=7.8cm]{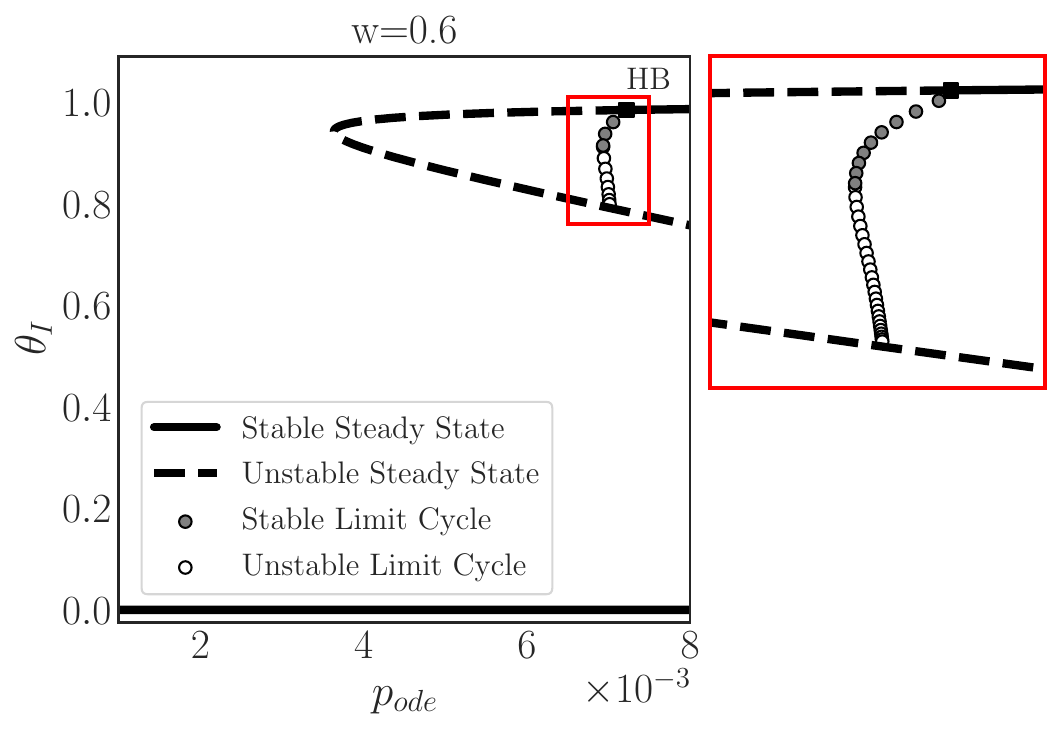}
\label{fig:Bifurcation Diagram a}
}
\subfigure[]{
       \includegraphics[width=5cm]{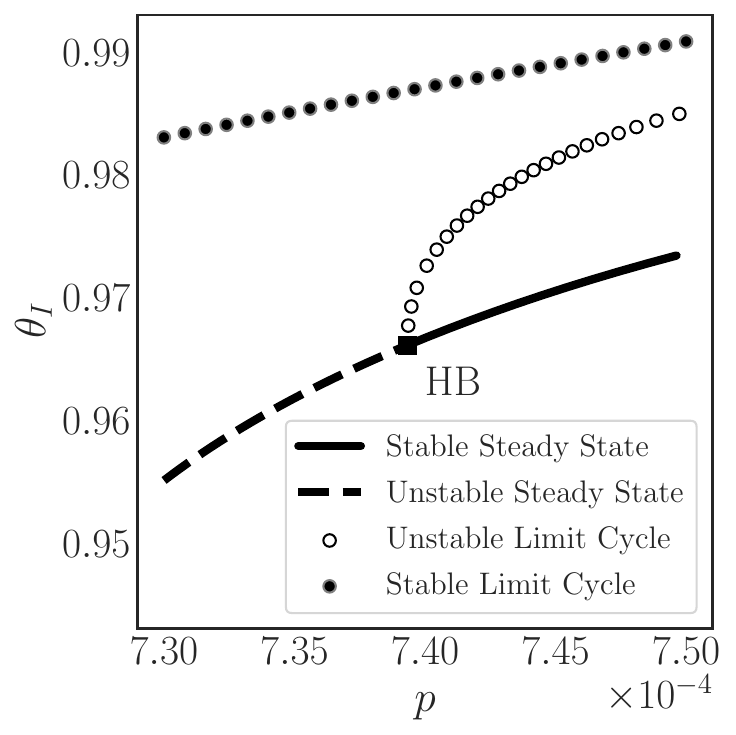}
       \label{fig:Bifurcation Diagram b}}
\caption{
(a) The bifurcation diagram of the mean-field SIS model (equation \eqref{eq:equations_Gross}) with respect to the infecteousness parameter $p$ is shown in terms of the fraction of $\theta_I$ for $w'=0.6$. For this system, a supercritical Hopf bifurcation at $p_{ode}^c=7.07 \times 10^{-3}$ occurs. The stable steady states are represented as solid black lines, the unstable steady states as dashed black lines, and the stable limit cycle as a black circle with gray edge color. (b) The bifurcation diagram constructed from the eSDE's drift component is shown again in terms of $\theta_I$. Parameter continuation of this eSDE model reveals a now subcritical Hopf bifurcation at $p^c=7.39 \times 10^{-4}$. The stable steady states are represented as black solid lines, the unstable steady states as dashed black lines, the stable limit cycle as black solid circles with gray edge color, and the unstable limit cycle as white circles with black edge color. }
\label{fig:Bifurcation Diagram}
\end{figure}

We start by constructing the bifurcation diagram for the mean-field SIS model discussed in section \ref{sec:coarse_grained_model} with the software AUTO \cite{scigma2023}. The bifurcation diagram for this low-dimensional SIS model is shown in figure \ref{fig:Bifurcation Diagram a} for $w' = 0.6$. In section \ref{sec:Bifurcation_Coarse_ODEs} we show a comparison of the bifurcation diagrams for the mean-field SIS model for $w'=\{0.2,0.4,0.6\}$ as well. From the constructed bifurcation diagram in figure \ref{fig:Bifurcation Diagram a} %[Second Row] 
a stable steady state (illustrated with a solid black line) exists for all the values of parameter $p_{ode}$ at $\theta_I = 0$. For large values of $p_{ode}$ another stable steady coexists with $\theta_I =0$. By performing continuation of this second steady state a supercritical Hopf bifurcation occurs at \hbox{$p^c_{ode} = 7.07 \times 10^{-3}$}. For parameter values larger and equal to $p^c_{ode} = 7.07 \times 10^{-3}$ the approach to the steady state is oscillatory; for parameter values $p_{ode}<p^c_{ode}$ the stability of the steady state changes and a stable limit cycle exists. We performed a limit cycle continuation for this model, tracking the period of the limit cycle. As the parameter $p_{ode}$ decreases, a turning point (saddle-node) of limit cycles is observed, and the period of the new, unstable limit cycle increases until it becomes numerically intractable (goes to infinity) suggesting that the unstable limit cycle \textit{dies} in a homoclinic (infinite period) bifurcation, involving a saddle-type mean-field steady state (known to arise when the $\theta_I =0$ steady state loses stability at high $p_{ode}$ values in a transcritical bifurcation).

For the drift component of the identified eSDE model, to pinpoint the value of the parameter at which the subcritical (hard) bifurcation occurs -that underpins the tipping point behavior- the software AUTO was again used. In this case, in contrast to the mean-field SIS model, we do not have a closed-form expression of the (deterministic) dynamics; instead we have a trained neural network that identifies the drift (and another for the diffusivity). To construct the bifurcation diagram of the identified drift we use the corresponding trained neural network. We import the drift network and, symbolically, we compute the Jacobian of the network. Symbolic computation of the Jacobian of the network (automatic differentiation is also possible) reduces to computing the derivative of the output with respect to the input by repeatedly applying the chain rule. To make sure the network is differentiable we used  \textit{tanh} activation functions that are continuous and differentiable. The use of continuous and differentiable activation functions is instrumental in performing parameter continuation.% To strengthen this argument we illustrate in Section \ref{sec:BF_Diagram_Relu_network} of the Appendix the bifurcation diagram constructed by using ReLU activation functions for the same data set.

The constructed bifurcation diagram of the drift component of the parameter-dependent eSDE is shown in Figure \ref{fig:Bifurcation Diagram b}. A subcritical Hopf Bifurcation occurs at the critical parameter value $p^c=7.39 \times 10^{-4}$. For parameter values smaller than $p^c$ an unstable steady-steady surrounded by a large amplitude limit cycle exists. At the critical parameter value $p^c$ we have the creation of an unstable limit cycle.  At $p^c$ the unstable limit cycle disappears in a subcritical (``hard'') Hopf bifurcation and the steady state changes stability \cite{strogatz2018nonlinear}. For values of $p\ge p^c$ we have the coexistence of the stable steady state and the two limit cycles (stable and unstable). The stable limit cycle is present for all parameter values at which we have sampled data. In our previous work \cite{gross2008robust}, we showed that a saddle-node bifurcation of cycles leads to the collision and annihilation of the two limit cycles for even larger values of $p$.
We do not pursue this further here, but we provide in section \ref{sec:Sampling} trajectories for larger values of $p$ that support the claim that the large amplitude limit cycles are not observed anymore.

\subsection{Data-driven coarse variables and an identified eSDE on those}
\label{sec:diffusion_maps_results}

As we showed in the previous two sections, when mean-field coarse variables for the SIS full network model are known \textit{apriori}(e.g. $\theta_I$ and $g_{SS}$) the construction of an eSDE model in terms of those becomes possible. In this section, we assume that those variables are not available, and illustrate how manifold learning can be used to \textit{extract} latent variables from the network behavior data themselves.
In our work, the manifold learning scheme Diffusion Maps was applied to sampled snapshot data from the full SIS network model. In section \ref{sec:diffusion_maps} we provide details for the implementation of the Diffusion Maps algorithm.
In our case, the Diffusion Maps algorithm was applied in a data set $\mathbf{S}$ sampled at the infectiousness parameter value $p=7.5 \times 10^{-4}$. Two \textit{non-harmonic} Diffusion Maps coordinates were found to parameterize the data $\phi_1,\phi_2$. The selection of those coordinates was achieved by the local linear regression algorithm proposed by Dsilva et al. \cite{dsilva2018parsimonious} through the \textit{datafold} Python package \cite{Lehmberg2020}. %The Diffusion Maps coordinates $\phi_1,\phi_2$ are shown on the left and middle plots of Figure \ref{fig:dmaps_idendified_dynamics}.

\begin{figure}[ht]
\centering
\includegraphics[width=14cm]{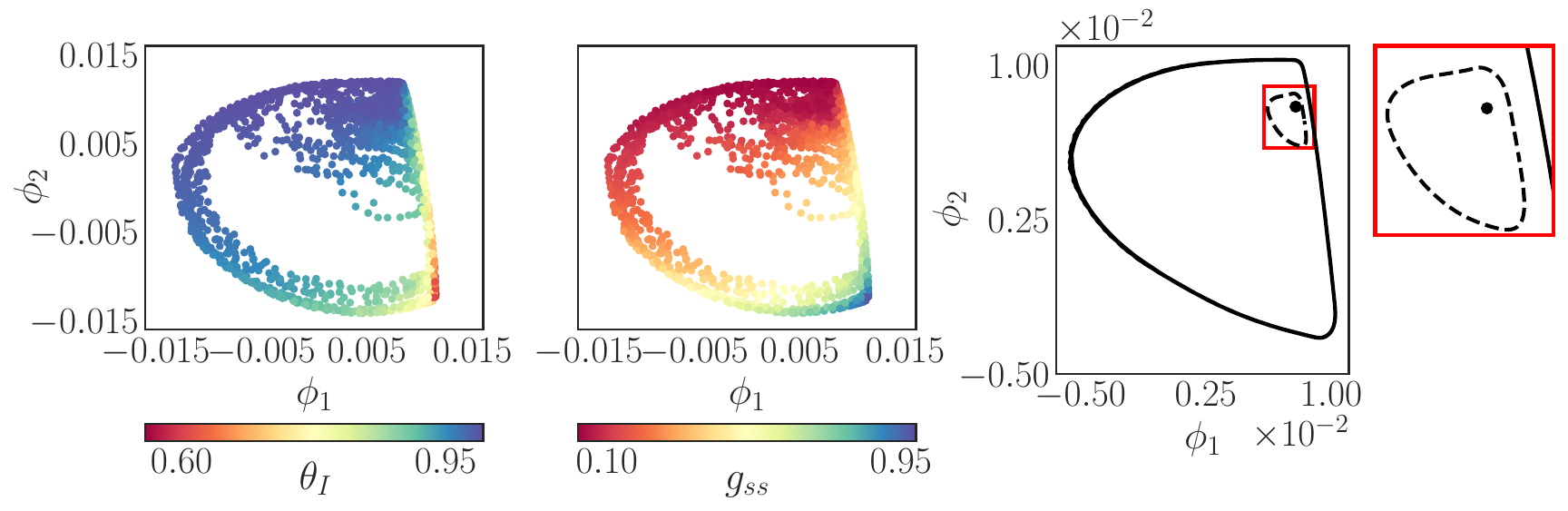}
\caption{The left and middle figures illustrate the two non-harmonic Diffusion Maps coordinates colored with the mean-field coarse variables $\theta_I$ and $g_{ss}$ indicating an approximate one-to-one correspondence. The right figure illustrates the identified stable  (solid black line) and unstable (dashed black line) limit cycles as well as the stable steady state (solid black circle) of the drift (deterministic component) of the eSDE in terms of the Diffusion Maps Coordinates $\phi_1,\phi_2$.} 
\label{fig:dmaps_idendified_dynamics}
\end{figure}

We provide a visual inspection of the \textit{interpretability} of the Diffusion Maps coordinates $\phi_1,\phi_2$ by coloring the mean-field coarse variables $\theta_I$ and $g_{SS}$ as functions of the Diffusion Maps coordinates; left and middle plots of figure \ref{fig:dmaps_idendified_dynamics}. A visual correlation appears, suggesting that the two physical coordinates $\theta_I$, $g_{SS}$ are functions of the Diffusion Maps coordinates $\phi_1,\phi_2$ as shown in figure \ref{fig:dmaps_idendified_dynamics} and \textit{vice-versa}.

Given the obtained Diffusion Maps coordinates $\phi_1,\phi_2$, we fit an eSDE with the neural network described in section \ref{sec:Learning_SDEs}. More details about the choice of hyperparameters and the model's architecture are provided in section \ref{sec:neural_network_details} of the Appendix. 

The drift dynamics (deterministic component) of the eSDE trained on the Diffusion Maps coordinates $\phi_1,\phi_2$ identifies a stable steady state surrounded by an unstable limit cycle. The unstable limit cycle is further surrounded by a stable limit cycle as shown on the right plot of figure \ref{fig:dmaps_idendified_dynamics}. The discovered deterministic attractor (based only on the drift of the eSDE) in the Diffusion Maps coordinates $\phi_1, \phi_2$ is in agreement with the attractor of the eSDE in terms of the mean-field variables $\theta_I$ and $g_{ss}$ (see section \ref{sec:learned_SDEs}). A more quantitative comparison between the two models and the full epidemiological model is provided in section \ref{sec:Mean_Escape_Time_results} where we estimate the corresponding mean escape times.

\subsection{Mean escape time computation}
\label{sec:Mean_Escape_Time_results}
Given the identified eSDEs at the infectiousness parameter value $p= 7.5 \times 10^{-4}$ in the mean-field coarse variables $\theta_I, g_{SS}$ as well as in the Diffusion Maps coordinates $\phi_1,\phi_2$, we compute the mean escape times for the two models (i) by using brute force stochastic simulations, (ii) by solving a stationary and (iii) an initial value BVP (see section \ref{sec:escape_times_methodology} for more details). We remind the reader that as \textit{escape time} we designate the time that is needed for a trajectory starting from the stable steady to \textit{escape} the boundary that is the unstable limit cycle for the first time. 

An estimation of the escape time for the full SIS network model was obtained by using brute-force simulations of the full model. The identified stationary state and the unstable limit cycle from the eSDE model on the mean-field coarse variables $\theta_I$,$g_{SS}$ were used to determine when a trajectory escaped. 

To initialize the simulations for the full SIS network model a \textit{nominal} graph was chosen. This nominal graph has values of $\theta_I$ and $g_{SS}$ similar to the identified stationary state of the drift part of the eSDE model at $p=7.5 \times 10^{-4}$. To obtain this nominal graph we simulate the full SIS network model and at each step, we compute $\theta_I$ and $g_{SS}$. When $\theta_I$ and $g_{SS}$ satisfy a termination criterion we stop the simulation and keep that graph as our nominal graph. The termination criterion chosen in our computations was that the $\ell^2$ norm between the identified deterministic steady state of the eSDE and the estimated $\theta_I$, $g_{SS}$ should be smaller than a threshold $\epsilon_t<  10^{-3}$. The mean escape time of the full SIS network model serves as the \textit{ground-truth} and offers a benchmark to evaluate the efficacy of the identified eSDEs in our two surrogate models.

It is worth mentioning that multiple epidemiological graphs might exist with the same $\theta_I$ and $g_{SS}$ but different topological connections. For the purpose of our escape time computations, reported below, we chose one representative graph that satisfies the criterion for $\theta_I$, $g_{SS}$ we mentioned above.

For the brute force stochastic simulations of the two eSDE surrogate models the numerical stochastic Euler-Maruyama scheme was used. The integration step was chosen to be $h=0.01$. The choice of the integration step was made based on our assumption that a step $h=0.01$ was sufficient for the identification of the eSDE's drift and diffusivity when training the neural network. Since a step $h=0.01$ is capable of identifying the dynamics is reasonable to assume that this is going to be a good step for estimating the escape time computations as well.  

\begin{figure}[ht]
\centering
\subfigure[]{
\includegraphics[height=4cm]{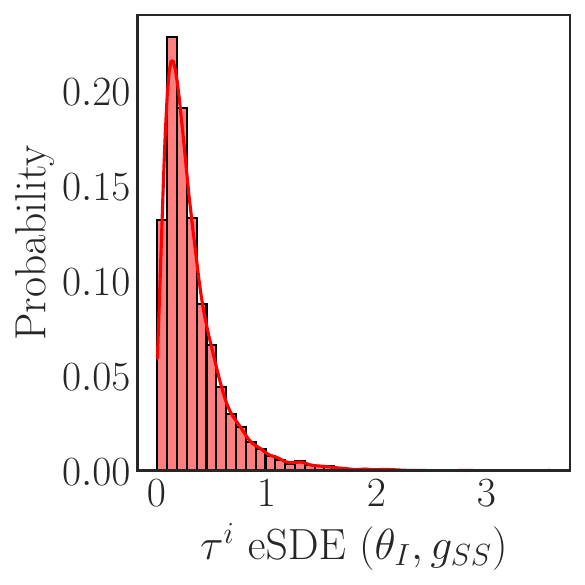}}
\subfigure[]{
\includegraphics[height=4cm]{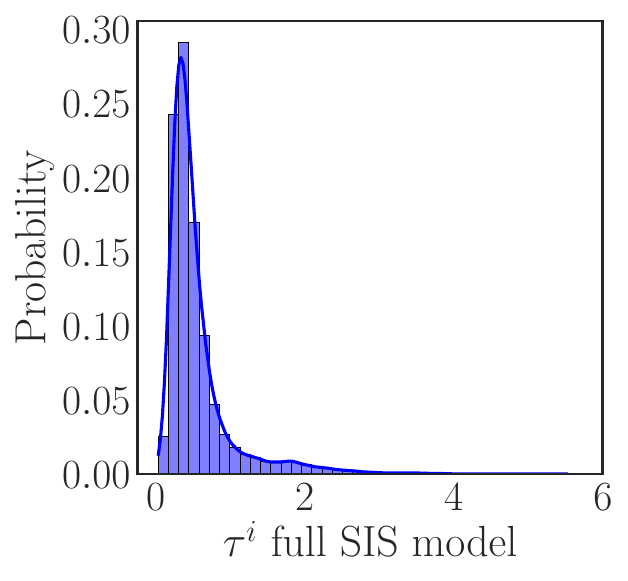}}
\\
\subfigure[]{
\includegraphics[height=4cm]{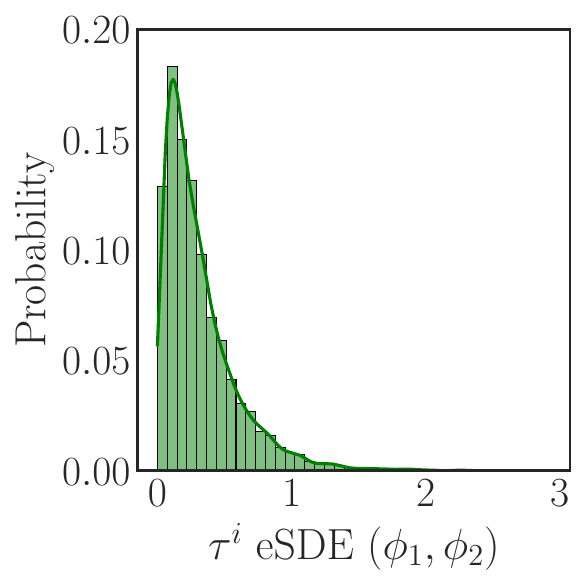}}
\subfigure[]{
\includegraphics[height=4cm]{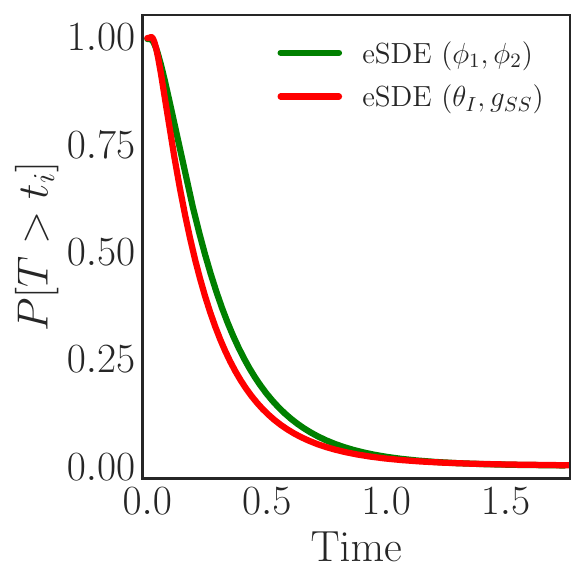}}
\caption{The obtained histograms from $10,000$ stochastic simulations and the approximate distributions with a kernel density estimation for: (a) the eSDE model on the mean-field variables ($\theta_I,g_{ss})$; (b) the full SIS network model; and (c) the eSDE model on the Diffusion Maps coordinates ($\phi_1,\phi_2$). (d) The probability of a trajectory that has not yet escaped through the boundary by a given time, $P[T>t_i]$, computed by solving the transient IBVP.} 
\label{fig:escape_time_distributions}
\end{figure}

For the brute force stochastic simulations we illustrate in figure \ref{fig:escape_time_distributions} the obtained histograms and the estimated distributions, obtained with kernel density estimation, for the: (a) eSDE model in terms of the mean-field coarse variables $\theta_I,g_{SS}$, (b) the full SIS network model, and (c) the eSDE model in terms of the Diffusion Maps coordinates $\phi_1,\phi_2$. For each of the three models, we simulated $10,000$ trajectories to estimate the mean escape times. In all three cases in figure \ref{fig:escape_time_distributions} the obtained distributions appear highly positively skewed, indicating a large standard deviation, which suggests an exponential distribution. The probability that a trajectory has not escaped the boundary at a given time is also shown in \ref{fig:escape_time_distributions}(d) for the two eSDEs estimated by solving the transient IBVP, see section \ref{sec:escape_times_methodology} and section \ref{sec:derivation_BVP_IBVP} of the Appendix.

In table \ref{tab:Mean_Escape_Comparison} we report the estimated values for the mean escape times with the brute force simulations for the three models. The mean escape times by solving the stationary BVP and the transient IBVP for the eSDE models are also included. For the brute force stochastic simulations the estimated standard deviation is reported too. The estimated standard deviations are in the same order of magnitude as the estimated means.

The largest discrepancy in the estimated escape times appears between the BVP computations for the eSDE model trained on the mean-field variables ($\theta_I,g_{SS}$) and the brute force computations for full SIS network model. The escape times estimated by solving the transient IBVP are smaller than the ones estimated by solving the stationary BVP. This is reasonable, since the solution of the transient IBVP is a truncated version of the full stationary problem.  We elaborate more in terms of the discrepancies between the estimated mean escape times between the surrogate eSDE models and the full SIS network model in the Discussion section \ref{sec:discussion_conclusion}. We also provide a more detailed description of the parameters used for the finite element computation with FEniCSx in section \ref{sec:Fenics_parameters} of the Appendix.

\begin{table}[ht]
    \centering
    \caption{Estimated mean escape times and standard deviations for the three different approaches.}
    \begin{tabular}{|c|c|c|c|}
        \hline 
         Approach & $\overline{\tau}$  eSDE $(\theta_I,g_{SS})$
         & $\overline{\tau}$ full SIS network model &  $\overline{\tau}$ eSDE $ (\phi_1,\phi_2)$ \\ \hline\hline
         Stochastic simulations & 0.334 (std 0.297) & 0.562 (std 0.475)
 & 0.317 (std 0.283)
         \\ \hline 
              Stationary BVP&  0.274 & - &  0.377
         \\ \hline 
              Transient IBVP  &  0.262 & - &  0.305
         \\ \hline 
    \end{tabular}
     \label{tab:Mean_Escape_Comparison}
\end{table}

\subsection{Computational Savings}
\label{sec:computational_savings}
In this section, we discuss the computational benefits of using the identified eSDEs, rather than the full SIS network model, for escape time computations. All the computations were performed on an Inter(R) Xeon (R) Silver 4112 CPU $@$ 2.60 GHz processor. To obtain an estimate of the computational time needed for each model we first estimate the time needed on average for a single function evaluation of each model.
 In table \ref{tab:function_evaluation}, the estimated time for a function evaluation is shown for the full SIS network model and for the two surrogate models. The times for a function evaluation for the two eSDEs are comparable, and both are $\sim$27 times smaller than the time needed for a function evaluation of the full SIS network model. 

The estimated average time for a single function evaluation was obtained by using $\sim$300 trajectories. Those simulated trajectories provide us with $\sim 10.000$ sample measurements of the time needed for each function evaluation.

\begin{table}[ht]
\centering
    \caption{Estimated function evaluation time for each model in seconds}
    \begin{tabular}{|c|c|c|}
    \hline
        eSDE ($\theta_I,g_{SS}$) & full SIS network model &  eSDE ($\phi_1,\phi_2$) \\ \hline\hline
         0.00326s & 0.0879s & 0.00330s 
         \\ \hline
    \end{tabular}
    \label{tab:function_evaluation}
\end{table}

By considering for each model: (a) the average time per function evaluation, (b) the total number of simulated trajectories, (c) and the average number of iterations per model to escape, we estimate the total computational time.  In table \ref{tab:Total_Time} we report the total time (in minutes) needed for the escape time computations for each of the three models and each of the two approaches we used. The escape time computations with stochastic simulations of the two eSDE models need $\sim$226 less time than those with the full SIS network model.

A quick explanation about why the ratio of the overall escape time calculations across models is an order of magnitude larger than the corresponding ratio for single function evaluations across the models. We remind the reader that for training the eSDE models we subsampled the trajectories of the full SIS network model and used snapshots for every 5 iterations of the full SIS model, assuming a step of $h=0.01$ between them. Therefore, a single step of the eSDE model corresponds to five steps of the full SIS network model. % This is also based on the assumption that by decreasing the dimensionality of the systems and constructing reduced-order dynamical eSDE models the stiffness of the dynamics also decreases and thus we are able to integrate with larger time steps.
Observe also that the escape time of the full SIS network model is $\sim1.7$ times longer than that of the eSDE models, and thus more simulation steps are required for a trajectory of the full SIS network model to escape.

The overall time needed for the mean escape time computations by solving the transient IBVP problem and the stationary BVP is even smaller than the one of the brute force simulations of the eSDEs. The eSDE trained in the mean-field coarse variables $\theta_I,g_{SS}$ needs about half a minute for the escape time computation by solving the stationary BVP, and the eSDE trained on the Diffusion Maps coordinate about 3 minutes. Note that solving BVPs and IBVPs and computing the mean escape time when the number of dimensions is small (e.g. smaller than 3-4) is computationally more efficient than using brute force simulations; in higher dimensions, however, this quickly becomes impractical and brute force simulations become the preferred approach.

 \begin{table}[ht]
\centering
    \caption{Wall clock time for escape time computations for each model in minutes, (i) with stochastic simulations, (ii) by solving the stationary BVP, and (iii) solving the transient IBVP.}
        \begin{tabular}{|c|c|c|c|}
        \hline 
         Approach & eSDE $(\theta_I,g_{SS})$
         & full SIS network model &  eSDE $ (\phi_1,\phi_2)$ \\ \hline\hline
         Stochastic simulations & 18.14 & 4116.65 & 15.85
         \\ \hline 
              Stationary BVP &  0.54 & - &  3.02
         \\ \hline 
              Transient IBVP &  5.03 & - &  13.77
         \\ \hline 
    \end{tabular}
    \label{tab:Total_Time}
\end{table}

For completeness, we mention that the time to train the neural networks for the eSDEs was $\sim 23$ minutes; thus the computational benefits of performing escape time computations with the surrogate eSDEs are still significant.

\section{Discussion - Conclusions}
\label{sec:discussion_conclusion}

We have presented a purely data-driven workflow for studying the tipping point dynamics of a complex SIS epidemiological adaptive network. This was accomplished by constructing reduced data-driven models in the forms of eSDEs for (i) the physical, coarse mean-field variables and  (ii) data-driven variables obtained with the manifold learning scheme Diffusion Maps. A deep learning scheme that identifies eSDEs directly from data (see section \ref{sec:learned_SDEs}) was used to approximately identify the dynamics in terms of the two alternative sets of coarse variables, and to disentangle the deterministic part of the dynamics from the stochastic part. Assisted by our parameter-dependent eSDE we find that a subcritical Hopf bifurcation occurs in the full SIS network model; we contrast this result with an analytical mean-field model proposed by Gross et al. \cite{gross2006epidemic}. We use the two eSDE models (in mean-field and in Diffusion Maps variables) to perform escape time computations, and compare the estimated mean values of the escape times of the eSDE models with the ones of the full SIS network model. We also discuss the computational savings for these computations by using the eSDE models for escape time computations rather than the full SIS network model.

Using data of the full SIS model collected when the infectiousness parameter $p$  takes the value $p=7.5 \times 10^{-4}$ for training, we showed that both our eSDE models, in physical as well as in data-driven coordinates, exhibit coexistence of a stable limit cycle surrounding an unstable limit cycle surrounding a stable steady state.
 From the performed escape time computations, see section \ref{sec:Mean_Escape_Time_results}, it was shown that the obtained mean escape times between the two eSDE surrogate models for the different approaches are in agreement. The estimated mean escape time of the full SIS network model is within the same order of magnitude, yet somewhat larger than the approaches we used for the eSDE models. This discrepancy may be due to an overestimation of the diffusivity in our eSDE models. To improve this discrepancy: (a) assuming a prior-partial structure of the dynamics (physics-infused) instead of fitting a black-box model might provide a better estimation of the mean escape times; (b) adding regularization/penalty terms for the diffusivity might also improve the estimation of mean escape times.  Despite the aforementioned discrepancy, our eSDE models were capable of providing a good qualitative description of the dynamics of the full SIS network model in terms of the two sets of coarse variables (mean-field and Diffusion Maps coordinates).

One limitation of our current approach is that our surrogate eSDE models do not provide detailed information about the full network's topology but rather a compressed summary of it. Therefore, a single point in our coarse variables (mean-field and diffusion maps) corresponds to a distribution of possible graphs. Being able to construct a map from the coarse variables back to the full graph, what is called \textit{lifting}, can become possible through generative models; recent approaches in the literature of generative models for network data such as NetGAN and \cite{bojchevski2018netgan} and adversarial graph autoencoders ARGA and ARVGA \cite{pan2018adversarially}) address this issue (see also \cite{crabtree2023gans} and, \cite{crabtree2023diffusion}). One possible extension could be to combine those generative models with our eSDE identification and learn simultaneously a latent space, the dynamics in this latent space, and a lifting map back to the original graph.

In this work, as the main example, we used an epidemiological model; yet our framework is easily transferable. Complex agent-based models inspired by applications in the financial market \cite{liu2015equation,fabiani2023tasks}, stochastic partial differential equations (SPDEs), \cite{prevot2007concise}, climate models \cite{gnanadesikan2018flux}, 
and lab experiments \cite{renson2019numerical,gonzalez2001characterization}
where tipping point dynamics occur, are cases where our proposed scheme is applicable to study tipping point behavior.

To identify our eSDE, we make a generic first choice of an it\'o-diffusion approximation by fitting the drift and diffusivity. This can be extended in the future by considering higher-order coefficients as well. Another possible future direction could be testing whether eSDEs 
 governed by a different stochastic process (e.g. $a-$stable L\'evy noise) can give more accurate eSDE models for the problem at hand. To this end, one could think of a \textit{cross-validation} approach where different type of processes are considered and compared.

We note that our methodology may be systematically extended to discover precursors of early warning signals from time-series data, including real world ones. Given the identified type of bifurcation in our best surrogate coarse grained model, characteristic patterns of fluctuations can be processed to construct early warnings of rare events \cite{scheffer2009early}. For
example, near a subcritical Hopf bifurcation like the one discovered here, as the critical eigenvalue pair approaches the imaginary axis zero, the impacts of physical perturbations along the corresponding eigenplane will decay slower and slower. This leads to an increased variance in the pattern of fluctuations observed. Their magnitude can provide an estimate
of the distance to and/or the probable time-interval until the next catastrophic event Other types of hard bifurcations,
e.g. saddle-nodes (turning points), can give rise to different types of catastrophic shifts and different relevant precursors.

% \textcolor{red}{If we can mathematically support a delay embedding of SDEs it might be an interesting point to make. The thought here is simple. Do a delay embedding of an SDE to identify the drift dynamics and do diffusion maps on the dynamics to check the \textit{true} dimensionality of the system. You can then learn again the SDE in a more parsimonious set of observables. As I mentioned above with Covid we did not measure the number of healthy individuals but just the number of cases (the number of infected) individuals but as we know one variable is not enough to have oscillations (limit cycles).  }

\section{Acknowledgements}
The authors would like to dedicate this paper to the memory of Professor Alexei Makeev, a long time friend and collaborator, a key contributor to the work, and an inspiring and knowledgeable scientist, whom we lost to Covid.
This work was partially supported by the US AFOSR FA9550-21-0317, the US Department of Energy SA22-0052-S001, and the National Science Foundation.

\bibliographystyle{unsrt}
\bibliography{lit.bib}

\begin{thebibliography}{10}

\bibitem{strogatz2018nonlinear}
Steven~H Strogatz.
\newblock {\em Nonlinear dynamics and chaos: with applications to physics,
  biology, chemistry, and engineering}.
\newblock CRC press, 2018.

\bibitem{guckenheimer2013nonlinear}
John Guckenheimer and Philip Holmes.
\newblock {\em Nonlinear oscillations, dynamical systems, and bifurcations of
  vector fields}, volume~42.
\newblock Springer Science \& Business Media, 2013.

\bibitem{uppal1974dynamic}
Ashok Uppal, Willis~Harmon Ray, and Aubrey~B Poore.
\newblock On the dynamic behavior of continuous stirred tank reactors.
\newblock {\em Chemical Engineering Science}, 29(4):967--985, 1974.

\bibitem{gross2006epidemic}
Thilo Gross, Carlos J~Dommar D’Lima, and Bernd Blasius.
\newblock Epidemic dynamics on an adaptive network.
\newblock {\em Physical review letters}, 96(20):208701, 2006.

\bibitem{gross2008robust}
Thilo Gross and Ioannis~G Kevrekidis.
\newblock Robust oscillations in sis epidemics on adaptive networks: Coarse
  graining by automated moment closure.
\newblock {\em Europhysics Letters}, 82(3):38004, 2008.

\bibitem{scheffer2020critical}
Marten Scheffer.
\newblock {\em Critical transitions in nature and society}, volume~16.
\newblock Princeton University Press, 2020.

\bibitem{gnanadesikan2018flux}
Anand Gnanadesikan, Richard Kelson, and Michaela Sten.
\newblock Flux correction and overturning stability: Insights from a dynamical
  box model.
\newblock {\em Journal of Climate}, 31(22):9335--9350, 2018.

\bibitem{gualdi2015tipping}
Stanislao Gualdi, Marco Tarzia, Francesco Zamponi, and Jean-Philippe Bouchaud.
\newblock Tipping points in macroeconomic agent-based models.
\newblock {\em Journal of Economic Dynamics and Control}, 50:29--61, 2015.

\bibitem{liu2015equation}
Ping Liu, CI~Siettos, C~William Gear, and IG~Kevrekidis.
\newblock Equation-free model reduction in agent-based computations:
  Coarse-grained bifurcation and variable-free rare event analysis.
\newblock {\em Mathematical Modelling of Natural Phenomena}, 10(3):71--90,
  2015.

\bibitem{omurtag2006modeling}
Ahmet Omurtag and Lawrence Sirovich.
\newblock Modeling a large population of traders: Mimesis and stability.
\newblock {\em Journal of Economic Behavior \& Organization}, 61(4):562--576,
  2006.

\bibitem{centola2018experimental}
Damon Centola, Joshua Becker, Devon Brackbill, and Andrea Baronchelli.
\newblock Experimental evidence for tipping points in social convention.
\newblock {\em Science}, 360(6393):1116--1119, 2018.

\bibitem{milkoreit2018defining}
Manjana Milkoreit, Jennifer Hodbod, Jacopo Baggio, Karina Benessaiah, Rafael
  Calder{\'o}n-Contreras, Jonathan~F Donges, Jean-Denis Mathias, Juan~Carlos
  Rocha, Michael Schoon, and Saskia~E Werners.
\newblock Defining tipping points for social-ecological systems
  scholarship—an interdisciplinary literature review.
\newblock {\em Environmental Research Letters}, 13(3):033005, 2018.

\bibitem{otto2020social}
Ilona~M Otto, Jonathan~F Donges, Roger Cremades, Avit Bhowmik, Richard~J
  Hewitt, Wolfgang Lucht, Johan Rockstr{\"o}m, Franziska Allerberger, Mark
  McCaffrey, Sylvanus~SP Doe, et~al.
\newblock Social tipping dynamics for stabilizing earth’s climate by 2050.
\newblock {\em Proceedings of the National Academy of Sciences},
  117(5):2354--2365, 2020.

\bibitem{gladwell2006tipping}
Malcolm Gladwell.
\newblock {\em The tipping point: How little things can make a big difference}.
\newblock Little, Brown, 2006.

\bibitem{dietrich2023learning}
Felix Dietrich, Alexei Makeev, George Kevrekidis, Nikolaos Evangelou, Tom
  Bertalan, Sebastian Reich, and Ioannis~G Kevrekidis.
\newblock Learning effective stochastic differential equations from microscopic
  simulations: Linking stochastic numerics to deep learning.
\newblock {\em Chaos: An Interdisciplinary Journal of Nonlinear Science},
  33(2):023121, 2023.

\bibitem{scigma2023}
scigma.
\newblock \url{https://github.com/scigma/scigma/tree/master/examples}, 2023.
\newblock Accessed: 2023-06-18.

\bibitem{coifman2006diffusion}
Ronald~R Coifman and St{\'e}phane Lafon.
\newblock Diffusion maps.
\newblock {\em Applied and computational harmonic analysis}, 21(1):5--30, 2006.

\bibitem{aggarwal2010managing}
Charu~C Aggarwal, Haixun Wang, et~al.
\newblock {\em Managing and mining graph data}, volume~40.
\newblock Springer, 2010.

\bibitem{bold2012equation}
Katherine~A Bold, Karthikeyan Rajendran, Bal{\'a}zs R{\'a}th, and Ioannis~G
  Kevrekidis.
\newblock An equation-free approach to coarse-graining the dynamics of
  networks.
\newblock {\em arXiv preprint arXiv:1202.5618}, 2012.

\bibitem{kattis2016modeling}
Assimakis~A Kattis, Alexander Holiday, Ana-Andreea Stoica, and Ioannis~G
  Kevrekidis.
\newblock Modeling epidemics on adaptively evolving networks: A data-mining
  perspective.
\newblock {\em Virulence}, 7(2):153--162, 2016.

\bibitem{Karthikeyan2017}
Karthikeyan Rajendran, Assimakis Kattis, Alexander Holiday, Risi Kondor, and
  Ioannis~G. Kevrekidis.
\newblock Data mining when each data point is a network.
\newblock In Pavel Gurevich, Juliette Hell, Bj{\"o}rn Sandstede, and Arnd
  Scheel, editors, {\em Patterns of Dynamics}, pages 289--317, Cham, 2017.
  Springer International Publishing.

\bibitem{wills2020metrics}
Peter Wills and Fran{\c{c}}ois~G Meyer.
\newblock Metrics for graph comparison: a practitioner’s guide.
\newblock {\em Plos one}, 15(2):e0228728, 2020.

\bibitem{athreya2022discovering}
Avanti Athreya, Zachary Lubberts, Youngser Park, and Carey~E Priebe.
\newblock Discovering underlying dynamics in time series of networks.
\newblock {\em arXiv preprint arXiv:2205.06877}, 2022.

\bibitem{oksendal2013stochastic}
Bernt Oksendal.
\newblock {\em Stochastic differential equations: an introduction with
  applications}.
\newblock Springer Science \& Business Media, 2013.

\bibitem{brunton2016discovering}
Steven~L Brunton, Joshua~L Proctor, and J~Nathan Kutz.
\newblock Discovering governing equations from data by sparse identification of
  nonlinear dynamical systems.
\newblock {\em Proceedings of the national academy of sciences},
  113(15):3932--3937, 2016.

\bibitem{chen2018neural}
Ricky~TQ Chen, Yulia Rubanova, Jesse Bettencourt, and David~K Duvenaud.
\newblock Neural ordinary differential equations.
\newblock {\em Advances in neural information processing systems}, 31, 2018.

\bibitem{NEURIPS2020_4a5876b4}
Patrick Kidger, James Morrill, James Foster, and Terry Lyons.
\newblock Neural controlled differential equations for irregular time series.
\newblock In H.~Larochelle, M.~Ranzato, R.~Hadsell, M.F. Balcan, and H.~Lin,
  editors, {\em Advances in Neural Information Processing Systems}, volume~33,
  pages 6696--6707. Curran Associates, Inc., 2020.

\bibitem{lu2019deeponet}
Lu~Lu, Pengzhan Jin, and George~Em Karniadakis.
\newblock Deeponet: Learning nonlinear operators for identifying differential
  equations based on the universal approximation theorem of operators.
\newblock {\em arXiv preprint arXiv:1910.03193}, 2019.

\bibitem{raissi2019physics}
Maziar Raissi, Paris Perdikaris, and George~E Karniadakis.
\newblock Physics-informed neural networks: A deep learning framework for
  solving forward and inverse problems involving nonlinear partial differential
  equations.
\newblock {\em Journal of Computational physics}, 378:686--707, 2019.

\bibitem{brandstetter2022clifford}
Johannes Brandstetter, Rianne van~den Berg, Max Welling, and Jayesh~K Gupta.
\newblock Clifford neural layers for pde modeling.
\newblock {\em arXiv preprint arXiv:2209.04934}, 2022.

\bibitem{kemeth2022learning}
Felix~P Kemeth, Tom Bertalan, Thomas Thiem, Felix Dietrich, Sung~Joon Moon,
  Carlo~R Laing, and Ioannis~G Kevrekidis.
\newblock Learning emergent partial differential equations in a learned
  emergent space.
\newblock {\em Nature Communications}, 13(1):3318, 2022.

\bibitem{lee2022learning}
Seungjoon Lee, Yorgos~M Psarellis, Constantinos~I Siettos, and Ioannis~G
  Kevrekidis.
\newblock Learning black-and gray-box chemotactic pdes/closures from agent
  based monte carlo simulation data.
\newblock {\em arXiv preprint arXiv:2205.13545}, 2022.

\bibitem{krischer1993model}
K~Krischer, R~Rico-Martinez, IG~Kevrekidis, HH~Rotermund, G~Ertl, and
  JL~Hudson.
\newblock Model identification of a spatiotemporally varying catalytic
  reaction.
\newblock {\em AIChE Journal}, 39(1):89--98, 1993.

\bibitem{rico1993discrete}
R~Rico-Martines, IG~Kevrekidis, MC~Kube, and JL~Hudson.
\newblock Discrete-vs. continuous-time nonlinear signal processing: Attractors,
  transitions and parallel implementation issues.
\newblock In {\em 1993 American Control Conference}, pages 1475--1479. IEEE,
  1993.

\bibitem{rico1992discrete}
Ramiro Rico-Martinez, K~Krischer, IG~Kevrekidis, MC~Kube, and JL~Hudson.
\newblock Discrete-vs. continuous-time nonlinear signal processing of cu
  electrodissolution data.
\newblock {\em Chemical Engineering Communications}, 118(1):25--48, 1992.

\bibitem{gonzalez1998identification}
Raul Gonz{\'a}lez-Garc{\'\i}a, Ramiro Rico-Mart{\`\i}nez, and Ioannis~G
  Kevrekidis.
\newblock Identification of distributed parameter systems: A neural net based
  approach.
\newblock {\em Computers \& chemical engineering}, 22:S965--S968, 1998.

\bibitem{li2020scalable}
Xuechen Li, Ting-Kam~Leonard Wong, Ricky~TQ Chen, and David~K Duvenaud.
\newblock Scalable gradients and variational inference for stochastic
  differential equations.
\newblock In {\em Symposium on Advances in Approximate Bayesian Inference},
  pages 1--28. PMLR, 2020.

\bibitem{yang2020physics}
Liu Yang, Dongkun Zhang, and George~Em Karniadakis.
\newblock Physics-informed generative adversarial networks for stochastic
  differential equations.
\newblock {\em SIAM Journal on Scientific Computing}, 42(1):A292--A317, 2020.

\bibitem{yang2022generative}
Liu Yang, Constantinos Daskalakis, and George~E Karniadakis.
\newblock Generative ensemble regression: Learning particle dynamics from
  observations of ensembles with physics-informed deep generative models.
\newblock {\em SIAM Journal on Scientific Computing}, 44(1):B80--B99, 2022.

\bibitem{kidger2021neural}
Patrick Kidger, James Foster, Xuechen Li, and Terry~J Lyons.
\newblock Neural sdes as infinite-dimensional gans.
\newblock In {\em International Conference on Machine Learning}, pages
  5453--5463. PMLR, 2021.

\bibitem{zhu2023learning}
Yuanran Zhu, Yu-Hang Tang, and Changho Kim.
\newblock Learning stochastic dynamics with statistics-informed neural network.
\newblock {\em Journal of Computational Physics}, 474:111819, 2023.

\bibitem{fang2022end}
Cheng Fang, Yubin Lu, Ting Gao, and Jinqiao Duan.
\newblock An end-to-end deep learning approach for extracting stochastic
  dynamical systems with $\alpha$-stable l{\'e}vy noise.
\newblock {\em Chaos: An Interdisciplinary Journal of Nonlinear Science},
  32(6):063112, 2022.

\bibitem{hasan2021identifying}
Ali Hasan, Jo{\~a}o~M Pereira, Sina Farsiu, and Vahid Tarokh.
\newblock Identifying latent stochastic differential equations.
\newblock {\em IEEE Transactions on Signal Processing}, 70:89--104, 2021.

\bibitem{pearson1901liii}
Karl Pearson.
\newblock Liii. on lines and planes of closest fit to systems of points in
  space.
\newblock {\em The London, Edinburgh, and Dublin philosophical magazine and
  journal of science}, 2(11):559--572, 1901.

\bibitem{roweis2000nonlinear}
Sam~T Roweis and Lawrence~K Saul.
\newblock Nonlinear dimensionality reduction by locally linear embedding.
\newblock {\em science}, 290(5500):2323--2326, 2000.

\bibitem{belkin2003laplacian}
Mikhail Belkin and Partha Niyogi.
\newblock Laplacian eigenmaps for dimensionality reduction and data
  representation.
\newblock {\em Neural computation}, 15(6):1373--1396, 2003.

\bibitem{balasubramanian2002isomap}
Mukund Balasubramanian and Eric~L Schwartz.
\newblock The isomap algorithm and topological stability.
\newblock {\em Science}, 295(5552):7--7, 2002.

\bibitem{van2008visualizing}
Laurens Van~der Maaten and Geoffrey Hinton.
\newblock Visualizing data using t-sne.
\newblock {\em Journal of machine learning research}, 9(11), 2008.

\bibitem{mcinnes2018umap}
Leland McInnes, John Healy, and James Melville.
\newblock Umap: Uniform manifold approximation and projection for dimension
  reduction.
\newblock {\em arXiv preprint arXiv:1802.03426}, 2018.

\bibitem{kramer1991nonlinear}
Mark~A Kramer.
\newblock Nonlinear principal component analysis using autoassociative neural
  networks.
\newblock {\em AIChE journal}, 37(2):233--243, 1991.

\bibitem{evangelou2022double}
Nikolaos Evangelou, Felix Dietrich, Eliodoro Chiavazzo, Daniel Lehmberg, Marina
  Meila, and Ioannis~G Kevrekidis.
\newblock Double diffusion maps and their latent harmonics for scientific
  computations in latent space.
\newblock {\em arXiv preprint arXiv:2204.12536}, 2022.

\bibitem{evangelou2021parameter}
Nikolaos Evangelou, Noah~J Wichrowski, George~A Kevrekidis, Felix Dietrich,
  Mahdi Kooshkbaghi, Sarah McFann, and Ioannis~G Kevrekidis.
\newblock On the parameter combinations that matter and on those that do not:
  data-driven studies of parameter (non) identifiability.
\newblock {\em PNAS Nexus}, 1(4):pgac154, 2022.

\bibitem{koronaki2023partial}
Eleni~D Koronaki, Nikolaos Evangelou, Yorgos~M Psarellis, Andreas~G Boudouvis,
  and Ioannis~G Kevrekidis.
\newblock From partial data to out-of-sample parameter and observation
  estimation with diffusion maps and geometric harmonics.
\newblock {\em arXiv preprint arXiv:2301.11728}, 2023.

\bibitem{talmon2013diffusion}
Ronen Talmon, Israel Cohen, Sharon Gannot, and Ronald~R Coifman.
\newblock Diffusion maps for signal processing: A deeper look at
  manifold-learning techniques based on kernels and graphs.
\newblock {\em IEEE signal processing magazine}, 30(4):75--86, 2013.

\bibitem{nadler2006diffusion}
Boaz Nadler, St{\'e}phane Lafon, Ronald~R Coifman, and Ioannis~G Kevrekidis.
\newblock Diffusion maps, spectral clustering and reaction coordinates of
  dynamical systems.
\newblock {\em Applied and Computational Harmonic Analysis}, 21(1):113--127,
  2006.

\bibitem{holiday2019manifold}
Alexander Holiday, Mahdi Kooshkbaghi, Juan~M Bello-Rivas, C~William Gear,
  Antonios Zagaris, and Ioannis~G Kevrekidis.
\newblock Manifold learning for parameter reduction.
\newblock {\em Journal of computational physics}, 392:419--431, 2019.

\bibitem{sroczynski2022questionnaires}
David~W Sroczynski, Felix~P Kemeth, Ronald~R Coifman, and Ioannis~G Kevrekidis.
\newblock Questionnaires to pdes: From disorganized data to emergent generative
  dynamic models.
\newblock {\em arXiv preprint arXiv:2204.11961}, 2022.

\bibitem{yair2017reconstruction}
Or~Yair, Ronen Talmon, Ronald~R Coifman, and Ioannis~G Kevrekidis.
\newblock Reconstruction of normal forms by learning informed observation
  geometries from data.
\newblock {\em Proceedings of the National Academy of Sciences},
  114(38):E7865--E7874, 2017.

\bibitem{chiavazzo2017intrinsic}
Eliodoro Chiavazzo, Roberto Covino, Ronald~R Coifman, C~William Gear,
  Anastasia~S Georgiou, Gerhard Hummer, and Ioannis~G Kevrekidis.
\newblock Intrinsic map dynamics exploration for uncharted effective
  free-energy landscapes.
\newblock {\em Proceedings of the National Academy of Sciences},
  114(28):E5494--E5503, 2017.

\bibitem{ferguson2010systematic}
Andrew~L Ferguson, Athanassios~Z Panagiotopoulos, Pablo~G Debenedetti, and
  Ioannis~G Kevrekidis.
\newblock Systematic determination of order parameters for chain dynamics using
  diffusion maps.
\newblock {\em Proceedings of the National Academy of Sciences},
  107(31):13597--13602, 2010.

\bibitem{evangelou2022learning}
Nikolaos Evangelou, Felix Dietrich, Juan~M Bello-Rivas, Alex~J Yeh, Rachel~S
  Hendley, Michael~A Bevan, and Ioannis~G Kevrekidis.
\newblock Learning effective sdes from brownian dynamic simulations of
  colloidal particles.
\newblock {\em Molecular Systems Design \& Engineering}, 2023.

\bibitem{kemeth2021initializing}
Felix~P Kemeth, Tom Bertalan, Nikolaos Evangelou, Tianqi Cui, Saurabh Malani,
  and Ioannis~G Kevrekidis.
\newblock Initializing lstm internal states via manifold learning.
\newblock {\em Chaos: An Interdisciplinary Journal of Nonlinear Science},
  31(9), 2021.

\bibitem{kevrekidis2003equation}
Ioannis~G Kevrekidis, C~William Gear, James~M Hyman, Panagiotis~G Kevrekidis,
  Olof Runborg, Constantinos Theodoropoulos, et~al.
\newblock Equation-free, coarse-grained multiscale computation: enabling
  microscopic simulators to perform system-level analysis.
\newblock {\em Commun. Math. Sci}, 1(4):715--762, 2003.

\bibitem{bury2021deep}
Thomas~M Bury, RI~Sujith, Induja Pavithran, Marten Scheffer, Timothy~M Lenton,
  Madhur Anand, and Chris~T Bauch.
\newblock Deep learning for early warning signals of tipping points.
\newblock {\em Proceedings of the National Academy of Sciences},
  118(39):e2106140118, 2021.

\bibitem{patel2023using}
Dhruvit Patel and Edward Ott.
\newblock Using machine learning to anticipate tipping points and extrapolate
  to post-tipping dynamics of non-stationary dynamical systems.
\newblock {\em Chaos: An Interdisciplinary Journal of Nonlinear Science},
  33(2):023143, 2023.

\bibitem{lim2020predicting}
Soon~Hoe Lim, Ludovico Theo~Giorgini, Woosok Moon, and John~S Wettlaufer.
\newblock Predicting critical transitions in multiscale dynamical systems using
  reservoir computing.
\newblock {\em Chaos: An Interdisciplinary Journal of Nonlinear Science},
  30(12):123126, 2020.

\bibitem{kong2021machine}
Ling-Wei Kong, Hua-Wei Fan, Celso Grebogi, and Ying-Cheng Lai.
\newblock Machine learning prediction of critical transition and system
  collapse.
\newblock {\em Physical Review Research}, 3(1):013090, 2021.

\bibitem{xiao2021predicting}
Rui Xiao, Ling-Wei Kong, Zhong-Kui Sun, and Ying-Cheng Lai.
\newblock Predicting amplitude death with machine learning.
\newblock {\em Physical Review E}, 104(1):014205, 2021.

\bibitem{galaris2022numerical}
Evangelos Galaris, Gianluca Fabiani, Ioannis Gallos, Ioannis Kevrekidis, and
  Constantinos Siettos.
\newblock Numerical bifurcation analysis of pdes from lattice boltzmann model
  simulations: a parsimonious machine learning approach.
\newblock {\em Journal of Scientific Computing}, 92(2):34, 2022.

\bibitem{fabiani2023tasks}
Gianluca Fabiani, Nikolaos Evangelou, Tianqi Cui, Juan~M Bello-Rivas,
  Cristina~P Martin-Linares, Constantinos Siettos, and Ioannis~G Kevrekidis.
\newblock Tasks makyth models: Machine learning assisted surrogates for tipping
  points.
\newblock {\em arXiv preprint arXiv:2309.14334}, 2023.

\bibitem{coifman2005geometric}
Ronald~R Coifman, Stephane Lafon, Ann~B Lee, Mauro Maggioni, Boaz Nadler,
  Frederick Warner, and Steven~W Zucker.
\newblock Geometric diffusions as a tool for harmonic analysis and structure
  definition of data: Diffusion maps.
\newblock {\em Proceedings of the national academy of sciences},
  102(21):7426--7431, 2005.

\bibitem{dsilva2018parsimonious}
Carmeline~J Dsilva, Ronen Talmon, Ronald~R Coifman, and Ioannis~G Kevrekidis.
\newblock Parsimonious representation of nonlinear dynamical systems through
  manifold learning: A chemotaxis case study.
\newblock {\em Applied and Computational Harmonic Analysis}, 44(3):759--773,
  2018.

\bibitem{gillies2013shapely}
Sean Gillies.
\newblock The shapely user manual.
\newblock {\em URL https://pypi. org/project/Shapely}, 2013.

\bibitem{logg2010dolfin}
Anders Logg and Garth~N Wells.
\newblock Dolfin: Automated finite element computing.
\newblock {\em ACM Transactions on Mathematical Software (TOMS)}, 37(2):1--28,
  2010.

\bibitem{logg2012automated}
Anders Logg, Kent-Andre Mardal, and Garth Wells.
\newblock {\em Automated solution of differential equations by the finite
  element method: The FEniCS book}, volume~84.
\newblock Springer Science \& Business Media, 2012.

\bibitem{Lehmberg2020}
Daniel Lehmberg, Felix Dietrich, Gerta K{\"o}ster, and Hans-Joachim Bungartz.
\newblock datafold: data-driven models for point clouds and time series on
  manifolds.
\newblock {\em Journal of Open Source Software}, 5(51):2283, 2020.

\bibitem{bojchevski2018netgan}
Aleksandar Bojchevski, Oleksandr Shchur, Daniel Z{\"u}gner, and Stephan
  G{\"u}nnemann.
\newblock Netgan: Generating graphs via random walks.
\newblock In {\em International conference on machine learning}, pages
  610--619. PMLR, 2018.

\bibitem{pan2018adversarially}
Shirui Pan, Ruiqi Hu, Guodong Long, Jing Jiang, Lina Yao, and Chengqi Zhang.
\newblock Adversarially regularized graph autoencoder for graph embedding.
\newblock {\em arXiv preprint arXiv:1802.04407}, 2018.

\bibitem{crabtree2023gans}
Ellis~R Crabtree, Juan~M Bello-Rivas, Andrew~L Ferguson, and Ioannis~G
  Kevrekidis.
\newblock Gans and closures: Micro-macro consistency in multiscale modeling.
\newblock {\em Multiscale Modeling \& Simulation}, 21(3):1122--1146, 2023.

\bibitem{crabtree2023diffusion}
Ellis~R Crabtree, Juan~M Bello-Rivas, and Ioannis~G Kevrekidis.
\newblock Micro-macro consistency in multiscale modeling: Score-based model
  assisted sampling of dynamical systems.
\newblock {\em In prepration}.

\bibitem{prevot2007concise}
Claudia Pr{\'e}v{\^o}t and Michael R{\"o}ckner.
\newblock {\em A concise course on stochastic partial differential equations},
  volume 1905.
\newblock Springer, 2007.

\bibitem{renson2019numerical}
Ludovic Renson, Jan Sieber, David~AW Barton, AD~Shaw, and SA~Neild.
\newblock Numerical continuation in nonlinear experiments using local gaussian
  process regression.
\newblock {\em Nonlinear Dynamics}, 98:2811--2826, 2019.

\bibitem{gonzalez2001characterization}
R.~Gonz{\'a}lez-Garc{\'i}a, R.~Rico-Mart{\'i}nez, Wilfried Wolf, Margot
  L{\"u}bke, Markus Eiswirth, Jason~S. Anderson, and Ioannis~G. Kevrekidis.
\newblock Characterization of a two-parameter mixed-mode electrochemical
  behavior regime using neural networks.
\newblock {\em Physica D: Nonlinear Phenomena}, 151(1):27--43, 2001.

\bibitem{scheffer2009early}
Marten Scheffer, Jordi Bascompte, William~A Brock, Victor Brovkin, Stephen~R
  Carpenter, Vasilis Dakos, Hermann Held, Egbert~H Van~Nes, Max Rietkerk, and
  George Sugihara.
\newblock Early-warning signals for critical transitions.
\newblock {\em Nature}, 461(7260):53--59, 2009.

\bibitem{risken1989}
H.~Risken.
\newblock {\em The {Fokker}-{Planck} {Equation}}, volume~18 of {\em Springer
  {Series} in {Synergetics}}.
\newblock Springer, 1989.

\bibitem{mshr2020}
B.~Kehlet, A.~Logg, J.~Ring, and G.~N. Wells.
\newblock mshr.
\newblock \url{https://bitbucket.org/fenics-project/mshr/}, 2020.

\end{thebibliography}

\clearpage
\appendix

\section{Appendix}

\subsection{Sampling}
\label{sec:Sampling}
\begin{figure}[ht]
    \centering
    \includegraphics[width=16cm]{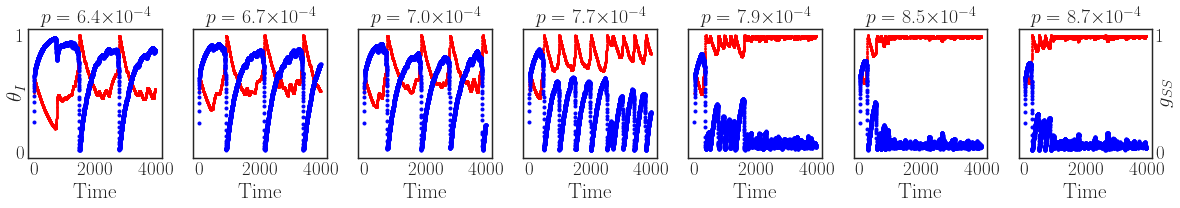}
   \includegraphics[width=16cm]{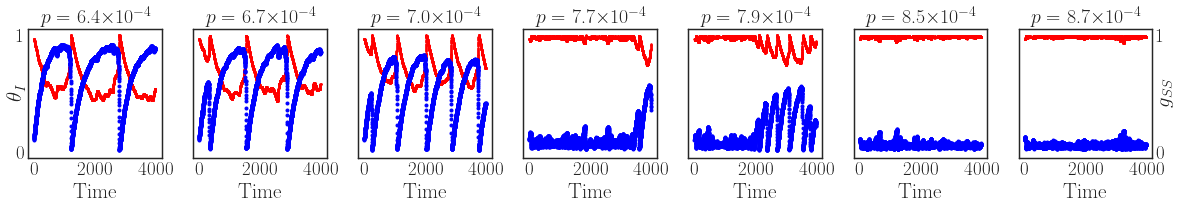}
\caption{For different values of the infectiousness parameter $p$ we illustrate sampled trajectories for two different initial conditions (initial graphs). The fraction of infected individuals $\theta_I$ is shown as blue points and the fraction of edges between susceptible individuals $g_{SS}$ is shown as red points.}
\label{fig:Mutiple_Trajectories}
\end{figure}

In this section, we report trajectories sampled across different values of the infectiousness parameter $p$ for the full SIS adaptive network model. In figure \ref{fig:Mutiple_Trajectories} for 7 values of the parameter $p$ we show two representative trajectories in terms of the mean-field variables $\theta_I$ and $g_{SS}$. As can be seen, from figure \ref{fig:Mutiple_Trajectories} large amplitude oscillations (suggesting the existence of an effective stable limit cycle) are present from $p=6.4 \times 10^{-4}$ until $p = 7.7 \times 10^{-4}$ (and perhaps $p = 7.9 \times 10^{-4}$). For larger values of $p$ (e.g. $p = 8.5 \times 10^{-4}$ and $p = 8.7 \times 10^{-4}$) those oscillations do not appear any more, and only an (effective) stationary steady state appears. Gross et al. \cite{gross2008robust} suggested that a saddle-node bifurcation of cycles takes place and this leads to the \textit{annihilation} of the deterministic mean-field limit-cycle. We do not pursue this further here.

\subsection{Bifurcation diagrams of the mean-field SIS model}
\label{sec:Bifurcation_Coarse_ODEs}

\begin{figure}[ht]
    \centering
    \includegraphics[width=15cm]{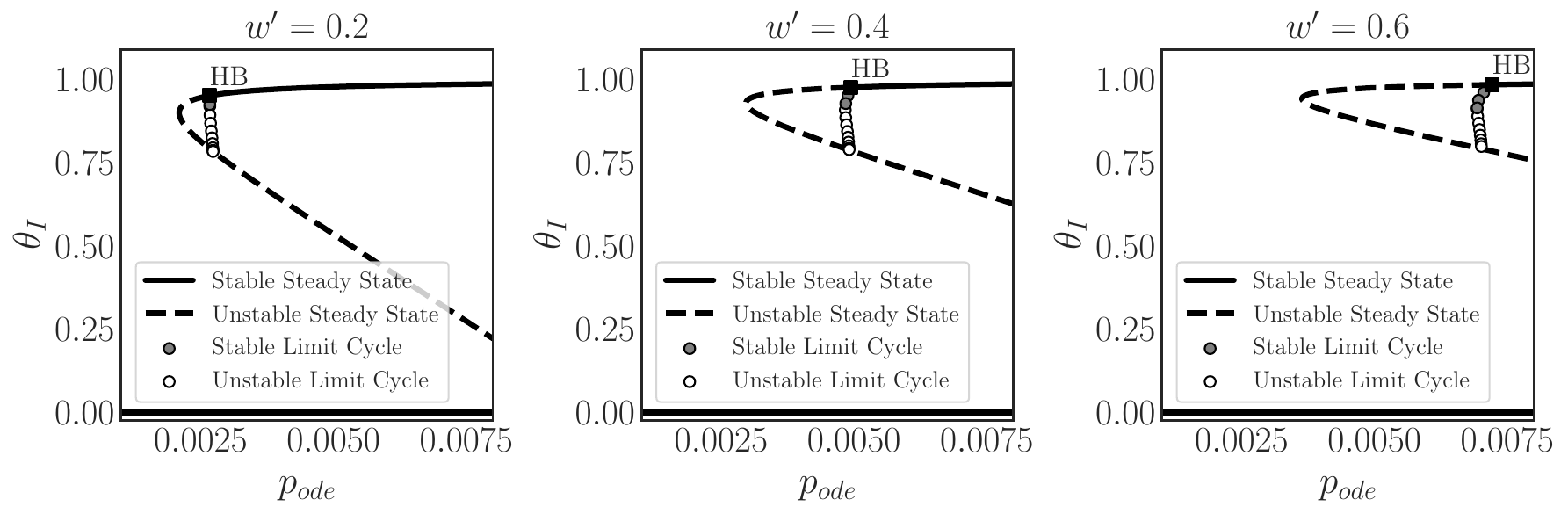}
    \caption{For three values of the  rewiring parameter $\vect{w}' = \{0.2,0.4,0.6\}$ respectively, the bifurcation diagrams w.r.t $p$ for the mean-field models are shown. For all three cases, a supercritical Hopf bifurcation, marked HB in the figure, occurs. }
\label{fig:bifurcation_diagrams_odes}
\end{figure}

For the mean-field SIS model described in section \ref{sec:Bifurcation_Coarse_ODEs} we constructed the bifurcation diagrams w.r.t. the infectiuousness parameter $p$ for three values of the rewiring parameter $w'$. In figure \ref{fig:bifurcation_diagrams_odes} we illustrate for $\vect{w}' = \{0.2,0.4,0.6\}$ that a stable (``healthy", epidemic extinction) steady state exists at $\theta_I=0$. For larger values of the parameter $p_{ode}$ --after a turning point bifurcation visible in the picture, and up to a transcritical bifurcation of the
extinction solution beyond the right bounday of the picture) another stable steady state (as well as another unstable one) coexists with the ``healthy" one at $\theta_I=0$. Continuation of this second stable steady state helps locate a supercritical Hopf Bifurcation point (HB in the figure) that occurs for all three values of $w'$. 

\subsection{Neural Network Architecture Details}
\label{sec:neural_network_details}
In this section, we provide a short description of the architecture and the hyperparameters chosen for the two neural networks trained to identify our eSDEs. For the parameter-dependent eSDE in terms of the mean-field coordinates ($\theta_I,g_{SS}$) we used $4$ layers and each layer had $25$ neurons. The batch size was selected as $30$ and the learning rate was $10^{-4}$. The activation functions for the drift network were selected to be \textit{tanh}, and for the diffusivity network, they were selected as \textit{softplus}, allowing the identified diffusivity to be positive semidefinite. 

For the eSDE model trained on the Diffusion Maps coordinates an architecture with $4$ layers and $25$ neurons each was used. For training this network the batch size was $32$ and the learning rate was $5 \times 10^{-4}$. The activation functions for the drift network were chosen to be \textit{relu}, and for the diffusivity they were selected again as \textit{softplus}. For this network, before training, the network outliers were removed with \textit{z-score}.

For both networks, a step size $h=0.01$ was selected. The data were split 90$|$10 into train$|$validation data. At inference time, we used generated (based on the drift) trajectories to check the discovered deterministic attractor of the system. Additionally, we employed stochastic trajectories for estimating the escape times of the eSDE models and contrasted those with the escape times of the full SIS network model.

\subsection{Escape Time Computation BVP and IBVP}
\label{sec:derivation_BVP_IBVP}
A brief description of the stationary BVP and the transient IBVP for escape time computations is provided in this section. Recall that Dynkin's formula~\cite{oksendal2013stochastic} tells us that
\begin{equation}
  \label{eq:dynkins-formula}
  \E[ u(\x_t) \mid \x_0 = \x ]
  =
  u(\x)
  +
  \E\left[
    \int_0^{\tau} (\A u)(\x_s) \, \d s \, \bigg| \, \x_0 = {\x}
  \right],
\end{equation}
where $u \colon \R^n \to \R$ is a smooth function and $\A$ is the generator of the It\=o diffusion Equation \eqref{eq:sde_langevin}, given by
\begin{equation*}
  \A u
  =
  \sum_{i = 1}^n \nu_i \frac{\partial u}{\partial x_i}
  +
  \frac{1}{2} \sum_{i = 1}^n \sum_{j = 1}^n (\sigma \sigma^\top)_{ij} \frac{\partial^2 u}{\partial x_i \partial x_j},
\end{equation*}
$\in \Omega$.
If $u$ is a solution of the boundary value problem
\begin{equation}
  \label{eq:boundary-value-problem}
  \left\{
    \begin{aligned}
      -&\A u = 1, & \text{in $\Omega$}, \\
      &u \equiv 0, & \text{on $\partial \Omega$},
    \end{aligned}
  \right.
\end{equation}
then \eqref{eq:dynkins-formula} becomes
\begin{equation*}
  0
  =
  u(\x)
  -
  \E\left[
    \int_0^{\tau} \d s \, \bigg| \, \x_0 = \x
  \right],
\end{equation*}
which implies that
\begin{equation*}
  \E[{\tau} \mid \x_0 = \x] = u(\x).
\end{equation*}

Alternatively, we can consider the adjoint problem to~\eqref{eq:boundary-value-problem}, given by
\begin{equation}
  \label{eq:poisson}
  \left\{
    \begin{aligned}
      -&\A^\star v = \delta_{\x}, & \text{in $\Omega$}, \\
       &v \equiv 0, & \text{on $\partial \Omega$,}
    \end{aligned}
  \right.
\end{equation}
where
\begin{equation*}
  \A^\star v
  \equiv
  -\sum_{i = 1}^n  \frac{\partial}{\partial x_i} 
    \nu_i u
  +
  \frac{1}{2} \sum_{i = 1}^n \sum_{j = 1}^n
  \frac{\partial^2}{\partial x_i \partial x_j}
    (\sigma \sigma^\top)_{ij} u,
\end{equation*}
and $\delta_{\x}$ denotes the Dirac delta distribution concentrated at $\x  \in \Omega$.
Multiplying the partial differential equation in~\eqref{eq:boundary-value-problem} by a function $v$ and integrating over $\Omega$, we see that
\begin{align*}
  0
  &=
    -\int_\Omega (\A u)(\z) \, v(\z) \, \d z
    -
    \int_\Omega v(\z) \, \d z \\
  &=
    -\int_\Omega u(\z) \, (\A^\star v)(\z) \, \d \z
    -
    \int_\Omega v(\z) \, \d \z \\
  &=
    \int_\Omega u(\z) \, \delta_{\x}(\z) \, \d \z
    -
    \int_\Omega v(\z) \, \d \z \\
  &=
    u(\x)
    -
    \int_\Omega v(\z) \, \d \z
\end{align*}
Therefore,
\begin{equation}
\overline{\tau}=\E[{\tau} \mid \x_0 = \x] = \int_\Omega v(\z) \, \d \z
\end{equation}
gives us another expression for the mean escape time by solving a (stationary) boundary value problem.

Now observe that the solution $p = p(\z, t)$ of the parabolic partial differential equation
\begin{equation}
  \label{eq:transition}
  \left\{
    \begin{aligned}
      &p_t = \A^\star p, & \text{in $\Omega \times (0, +\infty)$}, \\
      &p(\cdot, 0) = \delta_{\x}, & \text{in $\Omega \times \{ 0 \}$}, \\
      &p \equiv 0, & \text{on $\partial \Omega \times (0, +\infty)$,}
    \end{aligned}
  \right.
\end{equation}
is the density of the transition measure~\cite[Chapter 8]{risken1989} of $\x(t)$, provided that $t < \tau$.
We can construct the solution of~\eqref{eq:poisson} from the solution of~\eqref{eq:transition} as
\begin{equation}
  v(\z)
  =
  \int_0^\infty p(\z, t) \, \d t.
\end{equation}
Indeed,
\begin{equation}
\label{eq:BVP_transient}
  (\A^\star v)(\z)
  =
  \int_0^{\infty} \A^\star p(\z, t) \, \d t
  =
  \int_0^{\infty} \frac{\partial p}{\partial t}(\z, t) \, \d t
  =
  -\lim_{t \to 0^+} p(\z, t)
  =
  -\delta_{\x}(\z),
\end{equation}
where we have used that $\lim_{t \to \infty} p(\z, t) = 0$.
It then follows that
\begin{equation}
  \bar{\tau}
  =
  \int_\Omega v(\z) \, \d \z
  =
  \int_0^\infty
  P(t)
  \, \d t,
\end{equation}
where $P(t) = \int_\Omega p(\z, t) \, \d \z$ is the probability that $\tau > t$ for $t \ge 0$. Equation \eqref{eq:BVP_transient} provides another expression for computing mean escape times by solving a time-dependent initial-boundary value problem.
We validated these formulations through an analytically solvable problem, of Brownian motion on a unit circle with escape time $\tau = 0.5$ and confirmed that the brute force simulations, the transient IBVP and the BVP can provide accurate estimations.

\subsection{Details on finite element computations}
\label{sec:Fenics_parameters}
For the escape time computations involving the stationary BVP and the transient IBVP  with finite elements, the FEniCSx library was used \cite{logg2010dolfin,logg2012automated}. For solving the stationary BVP we used Lagrange elements of degree one and a mesh comprising $500$ cells generated by the software library mshr~\cite{mshr2020}. 

For solving the transient IBVP Lagrange elements of degree one were used and the mesh comprised $100$ cells. The time step for the integration was selected as $dt=0.001$. As the termination condition in this case we chose the mass being smaller than $10^{-3}$. Larger cell numbers, smaller time steps, and more strict termination criteria gave escape times with discrepancies only on the third decimal place from the ones reported in the main text. 

The codes used for the escape times computations are included in the public repository  \url{https://gitlab.com/nicolasevangelou/epidemics}.

\subsection{Codes}
The code used to generate the results for this paper can be accessed from the public repository and they will become available upon acceptance of the paper \url{https://gitlab.com/nicolasevangelou/epidemics}.
\end{document}